\documentclass[lettersize,journal]{IEEEtran}
\usepackage{amsmath,amsfonts}
\usepackage{algorithmic}
\usepackage{algorithm}
\usepackage{array}
\usepackage{textcomp}
\usepackage{stfloats}
\usepackage{url}
\usepackage{verbatim}
\usepackage{graphicx}
\usepackage{cite}
\usepackage{multirow,booktabs,subcaption}
\usepackage{color}
\usepackage{xcolor}
\usepackage{pifont}
\usepackage{xspace}
\usepackage{array,setspace}
\usepackage[normalem]{ulem}
\useunder{\uline}{\ul}{}
\usepackage[pagebackref,breaklinks,colorlinks]{hyperref}
\def\eg{\emph{e.g.}}
\def\ie{\emph{i.e.}}

\newcommand{\tesla}{\!\,\texttt{MLPBase\xspace}}
\newcommand{\teslam}{\!\,\texttt{MLP\xspace}}
\newcommand{\teslams}{\!\,\texttt{MLP-S\xspace}}
\newcommand{\cmark}{\ding{51}}%
\newcommand{\xmark}{\ding{55}}%
\newcommand{\lone}{{\large \ding{192}\xspace}}
\newcommand{\ltwo}{{\large \ding{193}\xspace}}

\newcommand{\fire}{{\includegraphics[scale=0.04]{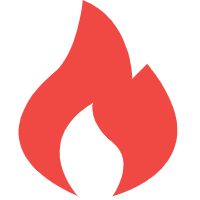}}}

\newcommand\blfootnote[1]{%
\begingroup 
\renewcommand\thefootnote{}\footnote{#1}%
\addtocounter{footnote}{-1}%
\endgroup 
}

\makeatletter
\def\endthebibliography{%
  \def\@noitemerr{\@latex@warning{Empty `thebibliography' environment}}%
  \endlist
}
\makeatother
\begin{document}

\title{MLP: Motion Label Prior for Temporal Sentence Localization in Untrimmed 3D Human Motions}

\author{Sheng Yan, Mengyuan Liu$^\dagger$,~\IEEEmembership{Member,~IEEE}, Yong Wang, Yang Liu, Chen Chen, Hong Liu,~\IEEEmembership{Member,~IEEE}}

\markboth{IEEE Transactions on Circuits and Systems for Video Technology}%
{MLP: Motion Label Prior for Temporal Sentence Localization in Untrimmed 3D Human Motions}

\maketitle


\blfootnote{$\dagger$ Corresponding author (Mengyuan Liu).}
\blfootnote{Sheng Yan and Yong Wang are with the School of Artificial Intelligence, Chongqing University of Technology, Chongqing 401120, China. (e-mail: eanson023@gmail.com; ywang@cqut.edu.cn).}
\blfootnote{Mengyuan Liu and Hong Liu are with the Key Laboratory of Machine
Perception, Shenzhen Graduate School, Peking University, Shenzhen 518055, China. (e-mail: nkliuyifang@gmail.com; hongliu@pku.edu.cn).}
\blfootnote{Yang Liu is with the College of Computer Science, Sichuan University, Chengdu 610065, China. (e-mail: liuyyy111@gmail.com)}
\blfootnote{Chen Chen is with the Center for Research in Computer
Vision, University of Central Florida, Florida 32826, USA. (e-mail: chen.chen@crcv.ucf.edu)}

\vspace{-8pt}
\begin{abstract}
In this paper, we address the unexplored question of temporal sentence localization in human motions (TSLM), aiming to locate a target moment from a 3D human motion that semantically corresponds to a text query. Considering that 3D human motions are captured using specialized motion capture devices, motions with only a few joints lack complex scene information like objects and lighting. Due to this character, motion data has low contextual richness and semantic ambiguity between frames, which limits the accuracy of predictions made by current video localization frameworks extended to TSLM to only a rough level. To refine this, we devise two novel label-prior-assisted training schemes: one embed prior knowledge of foreground and background to highlight the localization chances of target moments, and the other forces the originally rough predictions to overlap with the more accurate predictions obtained from the flipped start/end prior label sequences during recovery training. We show that injecting label-prior knowledge into the model is crucial for improving performance at high IoU. In our constructed TSLM benchmark, our model termed \teslam{} achieves a recall of 44.13 at IoU@0.7 on the BABEL dataset and 71.17 on HumanML3D (Restore), outperforming prior works. Finally, we showcase the potential of our approach in corpus-level moment retrieval. Our source code is openly accessible at \url{https://github.com/eanson023/mlp}.
\end{abstract}

\section{Introduction}
\label{sec:intro}

\begin{figure*}
    \centering
    \captionsetup{type=figure}
    \includegraphics[width=0.99\textwidth]{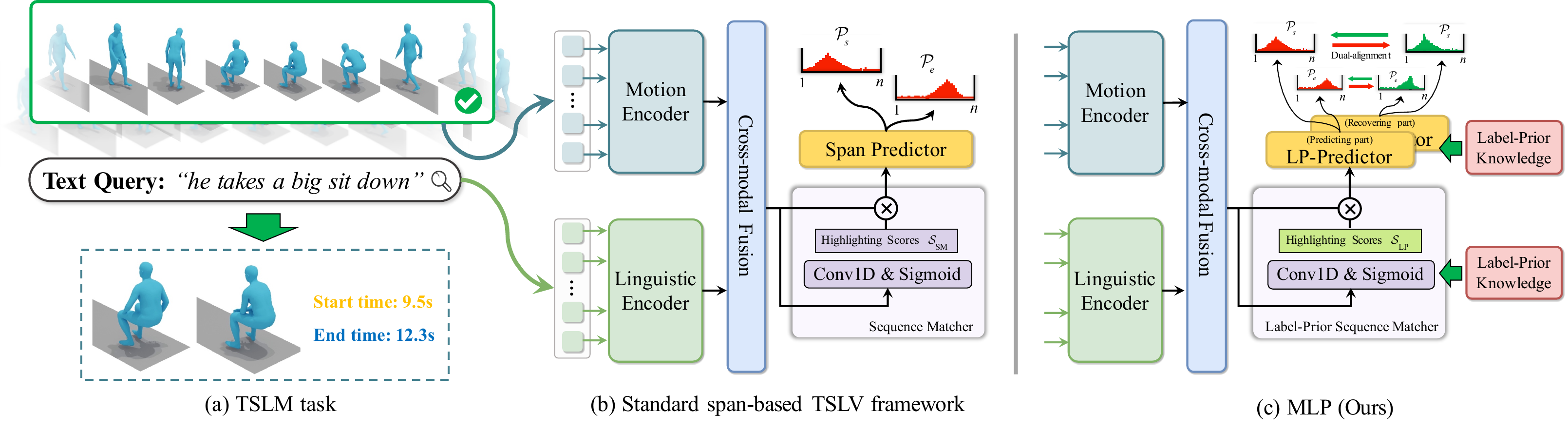}
    \vspace{-8pt}
    \captionof{figure}{\textbf{(a)} An illustration of the TSLM task. \textbf{(b)} Standard span-based TSLV framework~\cite{zhang2020span}, designed to predict the probability $\mathcal{P}_{s/e}$ for each motion frame as the starting/ending position of the target moment. \textbf{(c)} Our proposed \teslam{} injects label-prior knowledge into the two components of (b), corresponding to the Label-Prior Sequence Matcher (LP-Matcher) and the Label-Prior Span Predictor (LP-Predictor).}
    \vspace{-12pt}
    \label{fig:teaser}
\end{figure*}

Understanding what, when, and where a person is doing in a 3D world, is a fundamental and crucial question. Our goal is to retrieve or locate a \textit{target moment} in an untrimmed 3D human motion, given a text query describing the desired action (as illustrated in Fig.~\ref{fig:teaser}(a)). The ability to automatically match text descriptions with precise 3D human motion will open the door to numerous applications. For example, game developers with vast collections of 3D human motions can easily search for motions that align with specific text descriptions, enhancing virtual character animation.

In an effort to lower the cost of motion data capture, recent motion-text modeling primarily focuses on generating new motions conditioned on text~\cite{petrovich2022temos, jiang2023motiongpt, tevet2022human}. Simultaneously, some studies~\cite{yan2023cross,petrovich2023tmr,messina2023text} have investigated text-based retrieval of existing motions as a feasible alternative, realizing the suitability of reusing motions from a gallery for specific applications. These works involve responding to a given textual query and retrieving the most relevant motions from a \textbf{pre-segmented} or \textbf{trimmed} motion pool. They idealistically assume that motions have already been trimmed into semantically meaningful segments~\cite{guo2022generating,mandery2015kit}. Nevertheless, motion sequences in real-world scenario are frequently continuous and untrimmed~\cite{mahmood2019amass, punnakkal2021babel}, and the temporal information about when actions occur is not provided by their retrieval results. Motivated by this, we investigate the overlooked problem of \textcolor{purple}{t}emporal \textcolor{purple}{s}entence \textcolor{purple}{l}ocalization in untrimmed human \textcolor{purple}{m}otions (TSLM), which has not been explored to the best of our knowledge. 

To tackle TSLM, a straightforward approach is applying well-established temporal sentence localization in videos (TSLV) techniques~\cite{gao2017tall, zhang2020span, zhang2021parallel, gao2021learning, zhou2022thinking}, as motion is also evolving over time. We theoretically analyze potential bottlenecks in adapting prevalent TSLV methods for TSLM. These include techniques based on anchors, regressions, and spans.
\textbf{Anchor-based} methods~\cite{zhang2019man, yuan2019semantic, sun2023video, zhang2020learning, han2023momentum} excel in TSLV but are sensitive to manually crafted multi-scale anchors. In the motion dataset, BABEL \cite{punnakkal2021babel}, the relative length of target moments varies from 1\% to 99\%, making preset anchors may not cover all target moments. 
\textbf{Regression-based} methods~\cite{liu2022few, gao2022efficient, chen2022sagcn} assume a one-to-one relationship between queries and target moments by directly regressing start/end timestamps. For TSLM, however, motions often involve multiple repeated or similar moments, such as the ``raise arm" action in rehabilitation exercises. Multiple similar moments cannot be predicted with this method unless the network output is carefully designed.
Comparatively, \textbf{Span-based} methods~\cite{zhang2020span, zhang2021parallel, nawaz2022temporal} directly predict the probability of each motion frame being the start/end of the target moment (see Fig.~\ref{fig:teaser}(b)). Combining start and end probabilities for a motion with $n$ frames results in $n^{2}$ potential predictions, which covers all target moments and addresses the difficulty of predicting multiple similar moments.
Hence, we attempt to use span-based methods to address the TSLM task. One natural question arises: Do existing span-based TSLV methods work well on the TSLM?

\begin{figure}[t]
\centering
\includegraphics[width=0.96\columnwidth]{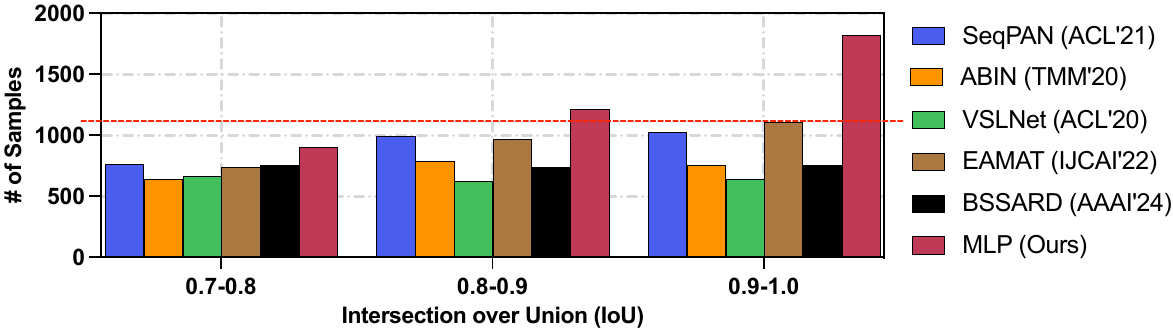}
\vspace{-5pt}
\caption{Comparison of performances between span-based TSLV method and ours at the high intersection over union (IoU) on BABEL dataset. (X-axis: different IoU thresholds; Y-axis: number of located samples)}
\label{fig:iou_compare}
\vspace{-0.5cm}
\end{figure}

\textbf{Problem Analysis:} We assessed recent span-based TSLV techniques on the BABEL dataset to address this question. As shown in Fig.~\ref{fig:iou_compare}, despite they generally perform well on IoU $\in [0.7, 0.9)$, only achieve inferior results at high IoU ($\geq 0.9$). To explain this, we carefully analyzed the gap when applying to TSLM: Unlike videos, 3D human motions are captured using specialized motion capture devices, with clean motions devoid of complex scene information like objects and lighting. If we consider `noun appearing after the verb in the query' as a proxy for scene information, \eg, \textit{``the man looks at his {\ul phone} then giggles"}, we can define the number of nouns as \textit{contextual richness} (recorded only the first occurrence). In the training set of HumanML3D (Restore) \cite{guo2022generating}, the contextual richness is only 1863, while this value is 4743 on the same word-count random subset of ActivityNet Captions \cite{krishna2017dense} used in TSLV. This discrepancy suggests that motions with fewer joints have less contextual richness, which also subtly reflects the semantic ambiguity between motion frames. As such, span-based TSLV methods only provide rough predictions.

\textbf{Our Solution:} Completing precise localization in human motions with ambiguous semantic differences poses a challenge. We wondered: Can we refine predictions even in scenarios with low contextual richness by informing the model with the start and end times of a query-motion pair beforehand? In contrast to existing methods, we considered injecting label-prior knowledge into the model. This is motivated by recent works that incorporate ground-truth information into models \cite{mihaylova2019scheduled, zhang2021parallel, zhang2022dino, li2022dn}. Two new components shown in Fig. \ref{fig:teaser}(c) correspond to the two novel label-prior-assisted training schemes that we propose --- the Label-Prior Sequence Matcher (\textbf{LP-Matcher}) and the Label-Prior Span Predictor (\textbf{LP-Predictor}):

\lone{} In LP-Matcher, target moments are treated as foreground and the rest as background. LP-Matcher embeds foreground-background labels into each motion frame during training. By using element-wise feature enhancement (Fig.~\ref{fig:teaser}(c)-bottom), the matched foreground regions are highlighted with the help of this prior knowledge, increasing the chances of locating the target moment. To address the training-inference discrepancy, we propose a simple yet effective \textit{Perturbation strategy} to mitigate it.
\ltwo{} For LP-Predictor (Fig.~\ref{fig:teaser}(c)-top), we cleverly come up with an additional label-prior-assisted training scheme. We divide LP-Predictor into two parts. In addition to the \textit{Predicting part}, which predicts the distribution of start/end boundaries of target moments normally, a \textit{Recovering part} is introduced in parallel. In this part, our goal is to recover the flipped start/end label sequences. 
Since these sequences are near to the ground truth, learning in this part is easier and the predicted distributions are more accurate.
Importantly, a \textit{Dual-alignment} approach is employed to enforce the Prediction part's distributions overlap with them. Fig. \ref{fig:iou_compare}-right demonstrates a significant improvement in high IoU with the injection of label-prior knowledge.

\textbf{Our contributions are the following:} 
(i) We investigate the unexplored problem of text-to-motion localization and construct a TSLM benchmark to evaluate the localization quality of comparable methods.
(ii) We propose \teslam{}, which stands for \textcolor{purple}{M}otion \textcolor{purple}{L}abel \textcolor{purple}{P}rior for Temporal Sentence Localization in Untrimmed 3D Human Motions.
Extensive experiments demonstrate that \teslam{}, with injected label-prior knowledge, achieving state-of-the-art performance.
(iii) We showcase the potential of \teslam{} in \textbf{corpus-level} moment retrieval, which aligns better with real-world application scenarios.
\section{Related Work}
\label{sec:related_work}

We present an overview of related work involving the use of text and motion modeling, and a brief discussion of the temporal localization in human motions. Finally, we conclude with related work on label-prior knowledge.
\medskip

\noindent\textbf{Text and Human Motion}. With the release of text-annotated datasets for large-scale motion capture collections such as HumanML3D \cite{guo2022generating} and KIT-ML \cite{mandery2015kit}, there has been a growing focus in the field of human motion modeling on bridging the semantic gap between natural language and 3D human motion. Particularly in text-conditioned motion generation, TEMOS \cite{petrovich2022temos} introduced a VAE model that learns a shared latent space for both motion and text. Following the trend of diffusion models \cite{ho2020denoising}, numerous works \cite{tevet2022human, zhang2022motiondiffuse, chen2023executing} have demonstrated motion generation based on diffusion. Leveraging the success of large language models (LLM), recent works \cite{jiang2023motiongpt, zhang2023motiongpt} introduced MotionGPT, treating multimodal signals as special input tokens within the LLM to generate continuous human motion sequences.

In addition to motion generation, some tasks have established bridges between human motion and natural language. MoLang \cite{kim2022learning}, by utilizing both paired and unpaired motion and language datasets, proposed a joint representation model with contrastive learning. Recently, some works \cite{yan2023cross, petrovich2023tmr, messina2023text} have similarly employed contrastive learning to explore text-to-motion retrieval, making up for the difficulty of generative models in generating authentic sequences. TEACH \cite{athanasiou2022teach}, an extension of TEMOS, synthesizes temporal motion compositions from a series of text descriptions. This can be seen as research in contrast to our work. Specifically, TEACH fundamentally synthesizes temporal motion compositions from input sequences (\ie, \{(duration, text description), (...)\}). In contrast, our work can be understood as estimating the duration required for each sub-motion that matches the text description from pre-existing temporal motion compositions, in direct contrast to TEACH.

\noindent\textbf{Temporal Localization in Human Motions}. Early works in motion or skeleton localization \cite{song2018spatio, wang2018beyond, sun2022locate, filtjens2022skeleton} trained neural models on datasets composed of pairs of motions and action labels \cite{liu2017pku, niemann2020lara}. For instance, convolutional network MS-GCN \cite{filtjens2022skeleton} and transformer model LocATE \cite{sun2022locate} aggregated each frame from untrimmed motion sequences to predict action label probabilities. Strictly speaking, these methods cannot be considered TSLM models as they are uni-modal. The located moments still require recognition based on a limited set of action labels, resembling a ``skeleton" version of object detection. More recently, NPose \cite{endo2023motion}, designed for HumanMotionQA, incorporates symbolic reasoning and modular design, introducing the concept of localization with temporal relationships. Most relevant to our work is TMR \cite{petrovich2023tmr}, which utilizes a pre-trained multimodal motion-text retrieval model to demonstrate zero-shot temporal localization based on textual descriptions. In our experiments, we benchmarked such methods on TSLM. While akin to these works in taking text as input and possessing a certain ``localization" capability, the key difference lies in our focus on supervised TSLM.

\noindent\textbf{Label-Prior Knowledge}. Many works leverage label-prior knowledge to endow models with a deeper understanding. For instance, classical neural machine translation (NMT) models \cite{vaswani2017attention, guo2022tm2t, sutskever2014sequence} are often trained using teacher forcing, where the model makes each decision based on the golden history of the target sequence (\ie, label-prior knowledge). Similarly, in the new paradigm of detection transformers (DETR) for object detection, many improvement works \cite{zhang2022dino, li2022dn, zong2023detrs} also utilize label-prior knowledge to speed up training convergence. They feed noise-added ground-truth labels and boxes into the decoder and train the model to reconstruct the original ones. Inspired by these works, our research proposes two novel label-prior-assisted training schemes to refine the rough predictions under the semantic ambiguity characteristic presented in human motion.
\section{Method}
\label{sec:method}

In this section, we first introduce the definition of the span-based TSLM task (Sec.~\ref{subsec:task_define})
Then, we provide a description of \teslam{} (Sec.~\ref{subsec:teslam}). 
Finally, we discuss training strategy (Sec.~\ref{subsec:training_strategy}).

\subsection{Task Definition} \label{subsec:task_define}

\noindent\textbf{Text query} refers to the description of the target moment, presented as written natural language sentences, such as \textit{``turning back to the left side"}. The data structure is a word sequence $\mathbf{Q} = (w_1, \ldots, w_N)$ from the English vocabulary, where each word is represented as $w \in \mathbb{R}^{d_w}$, and $N$ represents the length of the sequence, $d_w$ is the word embedding dimension.

\noindent\textbf{3D Human Motion} is defined as a sequence of 3D human poses, $\mathbf{M} = (p_{1}, \ldots, p_{F})$, where $F$ represents the number of time frames. Each pose ${p} \in {{\mathbb {R}}^{263}}$ corresponds to a representation of the articulated human body, including joint positions~\cite{holden2016deep}, rotations~\cite{zhou2019continuity}, foot concat, etc.
In cases of long motion sequences, they are segmented into $S$ interval snippets, and the average feature for each snippet is calculated. The segmented motion is then represented as $\mathbf{M} \in {{\mathbb {R}}^{T \times p}}$, where $T=min(F,S)$.

\noindent\textbf{Task Objective.} Given the motion $\mathbf{M}$ and the text query $\mathbf{Q}$, let $t_{s}$ and $t_{e}$ denote the starting and ending time points of ground-truth (target moment). Span-based TSLM aims to predict the start and end boundaries of the target moment. Consequently, it is necessary to map the starting/ending times $t_{s(e)}$ to the corresponding indices $i_{s(e)}$ in the motion sequence. Assuming the motion duration is $D$, the calculation formula for start/end indices $i_{s(e)}$ is as follows: $i_{s(e)} = \left\langle t_{s(e)}/D \times T \right\rangle$, where $\left\langle \cdot \right\rangle$ denotes the rounding operator. Given a pair $(\mathbf{M},\mathbf{Q})$ as input, span-based TSLM locates or predicts the time span from the start $i_{s}$ to the end $i_{e}$. In the inference process, the predicted span boundaries can be easily transformed into the corresponding time points via $t_{s(e)} = i_{s(e)}/T \times D$.

\begin{figure*}[t]
\centering
\includegraphics[width=1.0\textwidth]{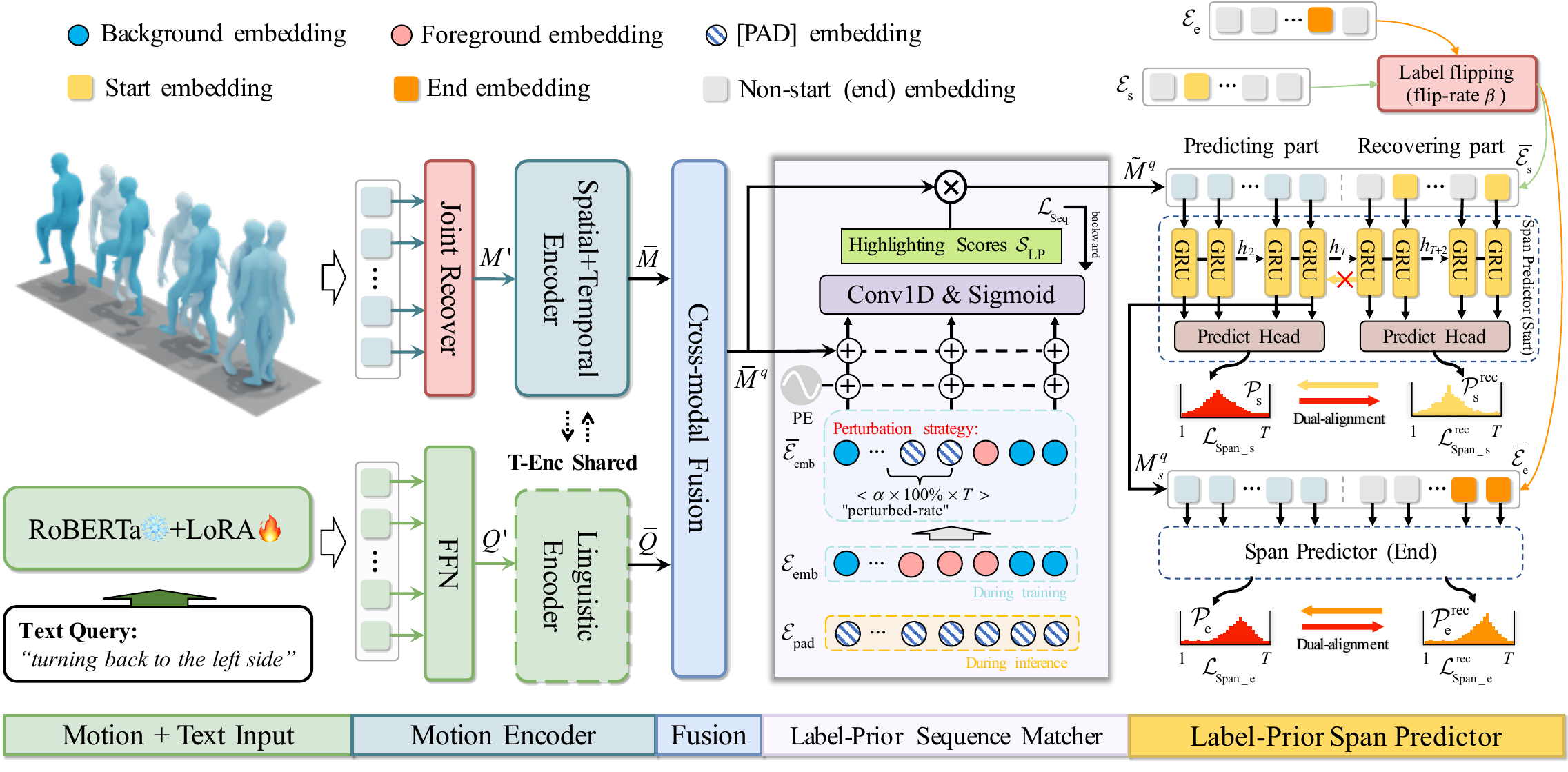}
\vspace{-15pt}
\caption{Our proposed \teslam{}: 
\textbf{(a)} For each $(\mathbf{M}, \mathbf{Q})$ pair, we build a factorised Spatial+Temporal Encoder to extract motion features $\mathbf{\bar{M}}$. Then, the Temporal Encoder (T-Enc) is shared to the linguistics modality to obtain text features $\mathbf{\bar{Q}}$. 
\textbf{(b)} Following Cross-modal Fusion, our Label-Prior Sequence Matcher embeds prior foreground-background knowledge ($\mathcal{E}_{\text{emb}}$) into the query-attended features $\mathbf{\bar{M}^{q}}$, which helps optimize $\mathcal{L}_{\text{Seq}}$ by pre-informs who is the foreground or background.
\textbf{(c)} In the Label-Prior Span Predictor, we design a Predicting part and a Recovering part in parallel. The optimization goal is the same for both parts; however, the Recovering part uses the label-flipped prior knowledge (composed of start/non-start or end/non-end label embeddings, $\mathcal{\bar{E}}_{\text{s/e}}$) to perform recovering training.
}
\label{fig:model_design}
\vspace{-0.5cm}
\end{figure*}

\subsection{MLP Model Architecture} \label{subsec:teslam}

As illustrated in Fig.~\ref{fig:model_design}, the \teslam{} model follows the ``\textcolor[rgb]{0.5058,0.7019,0.4}{En}\textcolor[rgb]{0.7215,0.3294,0.3137}{cod}\textcolor[rgb]{0.2745,0.5019,0.5529}{ing}-\textcolor[rgb]{0.4235,0.5568,0.7490}{Fusion}-\textcolor[rgb]{0.5333,0.4117,0.7333}{Match}-\textcolor[rgb]{0.8823,0.7568,0.3568}{Prediction}" paradigm. The details are explained below.

\medskip
\subsubsection{\textbf{Motion Encoder}} \label{subsubsec:mo_enc}

Our Motion Encoder is inspired by a variety of Factorised Spatio-Temporal models~\cite{arnab2021vivit, zhang2023towards, hu2023stdformer, liu2023transkeleton, zhu2022mlst}. Unlike previous approaches that treat motion as a pure temporal sequence~\cite{petrovich2022temos, guo2022generating, kim2023flame, tevet2022human}, we take a spatial-first feature extraction approach. Initially, we employ a Graph Convolutional Network (GCN) based spatial encoder~\cite{kipf2016semi} to introduce a hierarchical structure for motion encoding. The spatial-pooled features are then forwarded to an attention-based temporal encoder~\cite{vaswani2017attention} to capture interactions between poses at different time steps.

\noindent\textbf{Joint Recover}. With the hierarchical structure provided by the spatial encoder,  we first need to conduct structured representation recovery operations for the original poses. Each pose ${p}$ is transformed to $J=22$ joints, where the representation of each joint consists of a $d_{p}=12$-dim vector, redundantly encoding position, rotation, and velocity (including the root joint while omitting representations such as foot concatenation).

\noindent\textbf{Spatial Encoder (S-Enc).} The human body can be understood as a graph structure, with $G = \{\text{Nodes}, \text{Edges}\}$ representing the (joint, skeleton) connections within the body. $\mathcal{J} = (j_{1}, j_{2}, \ldots, j_{J})$ is used to represent nodes. The strength of edges in this graph, \ie, the correlation between joints, is represented by an adjacency matrix $\mathbf{A} \in \mathbb{R}^{J \times J}$. If joints are connected, we define the elements of the adjacency matrix as 1; otherwise, they are set to 0. The graph convolutional layer $l$ takes the matrix $\mathbf{H}^{(l)} \in \mathbb{R}^{J \times \lvert \mathbf{H} \rvert}$ as input, with $\lvert \mathbf{H} \rvert$ the number of features output by the previous layer. For example, for the first layer, the network takes the $J \times d_{p}$ matrix of motion elements\footnote{``Element" is a generic description that can refer to frames in motion sequences or processed snippets.} directly as input. Given this information and a set of trainable weights $\mathbf{W}^{(l)} \in \mathbb{R}^{\lvert \mathbf{H} \rvert \times d}$, the graph convolutional layer produces the following form of the matrix:
\begin{equation}
\mathbf{H}^{(l+1)}=BN\Big(tanh\big(\mathbf{A}^{(l)}\mathbf{H}^{(l)}\mathbf{W}^{(l)}\big)\Big)
\label{eq:gcn}
\end{equation}
where $\mathbf{A}^{(l)}$ is the adjacency matrix for layer $l$; $BN$ denotes batch normalization \cite{ioffe2015batch}. Following the standard deep learning form, we stack $L$ such layers to form the GCN. Subsequently, we employ mean pooling to aggregate spatial-aware motion feature, $\mathbf{M'} = MeanPool(\mathbf{H}_{1}^{(L)},\ldots,\mathbf{H}_{T}^{(L)})\in \mathbb {R}^{T \times d}$.

\noindent\textbf{Temporal Encoder (T-Enc).} Self-guided parallel attention (SGPA)~\cite{zhang2021parallel} is a transformer multi-modal variant that we use to model interactions between poses at various time positions. The SGPA simultaneously learn unimodal and cross-modal representations through two parallel multi-head attention blocks, which it then merges using a cross-gating strategy. For instance, in the first layer, the motion $\mathbf{M'}$ and text $\mathbf{Q'}$\footnote{The text has been dimension-aligned,  $\mathbf{Q'}=FFN(\mathbf{Q}) \in {{\mathbb {R}}^{N \times d}}$. We denote the single-layer feed-forward network as $FFN$  in this work.} are linearly projected to $\{Q_{M}^{(1)},K_{M}^{(1)},V_{M}^{(1)}\} \in {{\mathbb {R}}^{T \times d}}$ and $\{K_{Q}^{(1)},V_{Q}^{(1)}\} \in {{\mathbb {R}}^{N \times d}}$, respectively, the aggregation of new motion features $\mathbf{M}^{(2)}$ is defined as follows:
\begin{align}
    \mathbf{M}_{s}^{(1)} &= Softmax(\frac{Q_{M}^{(1)}{K_{M}^{(1)}}^{\text{T}}}{\sqrt{d}}) \times V_{M}^{(1)} \notag \\ 
    \mathbf{M}_{c}^{(1)} &= Softmax(\frac{Q_{M}^{(1)}{K_{Q}^{(1)}}^{\text{T}}}{\sqrt{d}}) \times V_{Q}^{(1)}  \notag \\
    \mathbf{M}^{(2)} &= Sigmoid \big(FFN(\mathbf{M}_{c}^{(1)})\big) \odot \mathbf{M}_{s}^{(1)} \notag \\
    &+ \text{ } Sigmoid \big(FFN(\mathbf{M}_{s}^{(1)})\big) \odot \mathbf{M}_{c}^{(1)}
    \label{eq:feature_enc}
\end{align}
where $\odot$ denotes the Hadamard product. In order to obtain the encoded features $\mathbf{\bar{M}} \in \mathbb{R}^{T \times d}$, T-Enc applies a stack of $N_{\text{SGPA}}$ blocks.

\noindent\textbf{Linguistic Encoder} has the same structure as the T-Enc. We find in our experiments that the encoding effectiveness is further enhanced when the parameters between the two modalities are shared. Thus, $\mathbf{\bar{Q}} \in \mathbb{R}^{N \times d}$ is the encoded feature obtained by switching the motion and text inputs.

\medskip
\subsubsection{\textbf{Cross-modal Fusion}}

We introduce context-query attention (CQA) \cite{seo2016bidirectional, yang2022entity,zhang2020span} to fuse the textual information for each motion element. 
CQA first calculates the similarity between each pair of features in $\mathbf{\bar{M}}$ and $\mathbf{\bar{Q}}$, producing $\mathcal{S} \in \mathbb{R}^{T \times N}$. Subsequently, it derives two sets of attention weights: $\mathcal{A}_{MQ} = \mathcal{S}_{r} \cdot \mathbf{\bar{Q}}^{\top}$ and $\mathcal{A}_{QM} = \mathcal{S}_{r} \cdot \mathcal{S}_{c}^{\top} \cdot \mathbf{\bar{M}}^{\top}$, where $\mathcal{S}_{r}$/$\mathcal{S}_{c}$ are Softmax-normalized along rows/columns of $\mathcal{S}$. The computation of query-aware motion features $\mathbf{M^{q}}$ is as follows:
\begin{equation}
\mathbf{M^{q}}=FFN\big([\mathbf{\bar{M}}; \mathcal{A}_{MQ}^{\top};\mathbf{\bar{M}} \odot \mathcal{A}_{MQ}^{\top}; \mathbf{\bar{M}} \odot 
 \mathcal{A}_{QM}^{\top} ]\big)
\end{equation}
where $\mathbf{M^{q}} \in \mathbb{R}^{T \times d}$, $[\cdot;\cdot]$ represents concatenation option. Then, the encoded features $\mathbf{\bar{Q}}$ from Linguistic Encoder are aggregated into a sentence representation $q \in \mathbb{R}^{d}$ with additive attention \cite{bahdanau2014neural}. This representation is connected to each element in $\mathbf{M^{q}}$ to $\mathbf{\bar{M}_{c}^{q}} = (\bar{p}_{1}^{t}, \ldots, \bar{p}_{T}^{t})$, where $\bar{p}_{i}^{t} = [p_{i}^{t}; q] \in \mathbb{R}^{2d}$. Lastly, the query-attended motion feature computation is given by $\mathbf{\bar{M}^{q}} = FFN(\mathbf{\bar{M}_{c}^{q}}) \in \mathbb{R}^{T \times d}$.

\medskip
\subsubsection{\textbf{Label-Prior Sequence Matcher (LP-Matcher)}} \label{subsubsec:lpmatcher}

Sequence matching, commonly employed as an auxiliary task for span-based methods \cite{yang2022entity, yang2022entity}, involves treating elements within the target moment as foreground and the rest as background. A simple Conv1D network with 1 kernel often serves as a binary classifier, predicting the confidence of each element belonging to the foreground or background:
\begin{equation}
    \mathcal{S}_{\text{SM}} = Sigmoid\big(Conv1D(\mathbf{\bar{M}^{q}})\big) \label{eq:highlighting}
\end{equation}
where the highlighting score $\mathcal{S}_\text{SM} \in \mathbb{R}^{T}$ represents the probability of each element belonging to the foreground or background.

However, as analyzed in the introduction, compared to videos, human motions exhibit low contextual richness, making it more challenging to match foreground-background between elements with ambiguous semantic differences. To this end, we designed the first novel label prior-assisted training scheme. The proposed LP-Matcher, as shown in Fig. \ref{fig:model_design}. During training, the input to the matcher additionally includes ground truth embedding sequences composed of foreground (background) labels. The score $\mathcal{S}_{\text{SM}}$ is replaced with a new highlighting score $\mathcal{S}_{\text{LP}}$, computed as follows:
\begin{equation}
    \mathcal{S}_{\text{LP}} = Sigmoid\big(Conv1D\big(\mathbf{\bar{M}^{q}} + \mathcal{E}_{\text{emb}} + \mathcal{E}_{\text{pos}}\big)\big) \label{eq:S_LP}
\end{equation}
where $\mathcal{E}_{\text{emb}} \in \mathbb{R}^{T \times d}$  represents the ground truth embedding sequences. This prior knowledge ``pre-informs" \teslam{} about who is foreground or background, leads to more accurate sequence matching; $\mathcal{E}_{\text{pos}} \in \mathbb{R}^{T \times d}$ is position encoding (PE) given in the form of sine functions. During inference, $\mathcal{E}_{\text{emb}}$ is replaced with embeddings $\mathcal{E}_{\text{pad}} \in \mathbb{R}^{T \times d}$ that contain $T$ \texttt{[PAD]} tokens, to prevent inference cheating.

With prior knowledge available, the model converges rapidly during training. However, due to the lack of such support during inference, the model's capacity to match sequence becomes extremely fragile.
To mitigate this discrepancy, we employ a simple yet effective \textbf{Perturbation strategy}. An availability mask is a binary matrix $mask \in \mathbb{R}^{T}$, containing $\left \langle \alpha \times 100\% \times T \right \rangle$ of $\mathbf{1}$(s).  A limited number of $\mathbf{1}$ are randomly (uniform) placed in continuous positions within the $mask$, representing the perturbed region of the sequence. Here, $\alpha$ is a hyperparameter denoting the \textit{perturb-rate}. Using this notation, perturbed label embeddings can be derived as:
\begin{equation}
    \mathcal{\bar{E}}_{\text{emb}} = mask \odot \mathcal{E}_{\text{pad}} + (1 - mask) \odot \mathcal{E}_{\text{emb}}
\label{eq:disturbed_E}
\end{equation}
Moreover, $\alpha$ linearly increases during training steps based on a pre-defined function, gradually reducing dependence on prior knowledge.
In this way, the $\mathcal{E}_{\text{emb}}$ in Eq.~\eqref{eq:S_LP} is updated to $\mathcal{\bar{E}}_{\text{emb}}$. The loss function for LP-Matcher is formulated as:
\begin{equation}
    \mathcal{L}_{\text{Seq}}=f_{\text{CE}}(\mathcal{S}_{\text{LP}},\mathcal{Y}_{h}) \label{eq:L_seq}
\end{equation}
where $f_{\text{CE}}$ denotes the cross-entropy loss function. $\mathcal{Y}_{h}$ is a ground-truth sequence consisting of 0 for background and 1 for foreground. The highlighted features $\mathbf{\widetilde{M}}$ are derived through feature weighting, given by $\mathbf{\widetilde{M}^{q}} = \mathbf{\mathcal{S}_{\text{LP}} \cdot \bar{M}^{q}}$.

\medskip
\subsubsection{\textbf{Label-Prior Span Predictor (LP-Predictor)}}
\label{subsubsec:lppredictor}

The span prediction is executed last. We design another novel label prior-assisted training scheme. Specifically, we divide LP-Predictor into two parts: \textit{Predicting part} and \textit{Recovering part}. In the first part, the predictor predicts the distribution of start/end boundaries normally. In the second part, we train to recover the flipped start/end label prior sequences. Then, we use a \textit{Dual-alignment} method to force the predicted distributions of the Predicting part to overlap with them. Below, we will detail the prediction of start boundaries as an example.

\noindent\textbf{Predicting part.} The span predictor (start) consists of a unidirectional $GRU$ and a prediction head, where the head is a 2-layer $FFN$ network with $LayerNorm$ and $ReLU$ activation. The score for the start boundary is computed as follows:
\begin{equation}
\mathcal{S}_{s} = PredHead(\mathbf{M_{s}^{q}}), \; \mathbf{M_{s}^{q}} = GRU(\mathbf{\widetilde{M}^{q}})
\end{equation}
 where $\mathcal{S}_{s} \in \mathbb{R}^{T}$, and then the probability distribution for start boundaries is calculated by $\mathcal{P}_{s} = Softmax(\mathcal{S}_{s})$. The training objective is:
\begin{equation}
    \mathcal{L}_{\text{Span\_s}}= f_{\text{CE}}(\mathcal{P}_{s}, \mathcal{Y}_{s}) \label{eq:L_span}
\end{equation}
where $\mathcal{Y}_{s}$ represents the labels for the starting boundary.

\noindent\textbf{Recovering part.} In this part, we first embed the start label sequences $\mathcal{Y}_{s} \to \mathcal{E}_{s} \in \mathbb{R}^{T \times d}$. For these sequences, we apply \textit{Label flipping}, meaning that we randomly swap start labels with non-start labels within the sequence. We have a hyperparameter $\beta$ to control the rate of flipping, and the flipped sequence can be represented as $\mathcal{\bar{E}}_{\text{s}}$. Subsequently, we concatenate it with the highlighted feature $\mathbf{\widetilde{M}^{q}}$ and fed to the span predictor (start) together, as illustrated in Fig.~\ref{fig:model_design} right. Our goal is to recover $\mathcal{\bar{E}}_{\text{s}}$ back to $\mathcal{E}_{s}$ by comprehending the hidden state $h_{T}$ passed by the $GRU$ (\ie, sequence-level highlighted feature). Consequently, a cross-entropy loss is used to optimize the recovery process:
\begin{equation}
    \mathcal{L}_{\text{Span\_s}}^{\text{rec}}= f_{\text{CE}}(\mathcal{P}_{\text{s}}^{\text{rec}}, \mathcal{Y}_{s}) \label{eq:L_span_rec}
\end{equation}
where $\mathcal{P}_{\text{s}}^{\text{rec}} \in \mathbb{R}^{T}$ is the predicted start boundary distribution for this part, obtained by softmaxing the scores. Intuitively, the forward process of the Recovering part is parallel to the Predicting part

Note that the $GRU$ is set to be unidirectional, ensuring that the Recovering part does not leak prior knowledge to the Predicting part.

\noindent\textbf{Dual-alignment}. 
In contrast to the Predicting part, the learning in the Recovering part is relatively easier, and the predicted distribution is more accurate because we only slightly flip the label sequences. To enforce the predictions of the Predicting part to be as close as possible to the Recovering part, we minimize the Kullback-Leibler (KL) divergence between the distributions $\mathcal{P}_{s}$ and $\mathcal{P}_{\text{s}}^{\text{rec}}$:
\begin{equation}
    \mathcal{L}_{\text{Align}}= D_{\text{KL}}(\mathcal{P}_{\text{s}}|| \mathcal{P}_{\text{s}}^{\text{rec}})
\end{equation}

Similarly, this is done for the prediction of the end boundary, with the only difference being conditioned on the start boundary (as shown in Fig.~\ref{fig:model_design} right-bottom).

\subsection{Training Strategy} \label{subsec:training_strategy}

The overall training loss for the model is defined as:
\begin{equation}
\mathcal{L} =\lambda_{1}\mathcal{L}_{\text{Seq}}+\lambda_{2}\mathcal{L}_{\text{Span}}+\lambda_{3}\mathcal{L}_{\text{Span}}^{\text{rec}}+\lambda_{4}\mathcal{L}_{\text{Align}}
\end{equation}
where $\mathcal{L}_{\text{Span}}=\frac{1}{2}\times (\mathcal{L}_{\text{Span\_s}} + \mathcal{L}_{\text{Span\_e}} )$ and $\mathcal{L}_{\text{Span}}^{\text{rec}}=\frac{1}{2}\times(\mathcal{L}_{\text{Span\_s}}^{\text{rec}} + \mathcal{L}_{\text{Span\_e}}^{\text{rec}})$. $\lambda_{1-4}$ are trade-off hyperparameters, set to 5.0, 1.0, 1.0, and 1.0 respectively. During the inference process, the predicted starting and ending boundaries $(\hat{i}_{s}, \hat{i}_{e})$ for a given motion-query pair are generated by maximizing the joint probability:
\begin{equation}
\begin{aligned}
    (\hat{{i}}_{s},\hat{{i}}_{e}) &=  \arg\max_{\hat{{a}}_{s},\hat{{a}}_{e}} \mathcal{P}_s(\hat{{a}}_{s})\times\mathcal{P}_e(\hat{{a}}_{e}) \\
    \text{p}^{se} &= \mathcal{P}_s(\hat{{i}}_{s})\times\mathcal{P}_e(\hat{{i}}_{e})
\end{aligned}
\label{eq:infer}
\end{equation}
where $\text{p}^{se}$ is the located score of the predicted boundaries. Here, using the transformation mentioned earlier (see Sec.~\ref{subsec:task_define}), $\hat{t}_{s(e)} = \hat{i}_{s(e)}/T \times D$, can be transformed into the final predicted starting/ending times. 
\section{Experiments}

\begin{table*}[t]
    \small
    \setlength{\tabcolsep}{2.2pt}
	\centering
	\begin{tabular}{ l c c r r r r r }
		\toprule
		Dataset  & \# Motions (train/val/test) & \# Text queries & $N_{\textrm{vocab}}$ & $\bar{L}_{motion}$ & $\bar{L}_{query}$ & $\bar{L}_{moment}$ & $\Delta_{moment}$  \\
		\midrule
        BABEL  & $3,984/-/1,356$ & $25,810/-/9,697$ & $1,772$ & $29.79s$ & $2.37$ & $2.45s$ & $4.33s$ \\
        \midrule
        HumanML3D (Restore)  & $4,628/56/1,100$ & $26,722/320/6,560$ & $5,091$ & $28.87s$ & $15.12$ & $9.42s$ & $1.50s$ \\
        \bottomrule
	\end{tabular}
        \vspace{-2pt}
	\caption{\small Statistics of TSLM datasets, where $N_{\textrm{vocab}}$ is vocabulary size of lowercase words, $\bar{L}_{motion}$ denotes average duration of motions, $\bar{L}_{query}$ denotes average number of words in sentence query, $\bar{L}_{moment}$ is average length of target moments in seconds, and $\Delta_{moment}$ is the standard deviation of target moment.}
        \vspace{-5pt}
	\label{tab:datatset}
\end{table*}

\begin{figure*}[t]
\centering
\includegraphics[width=0.99\linewidth,trim=0 5 0 5,clip]{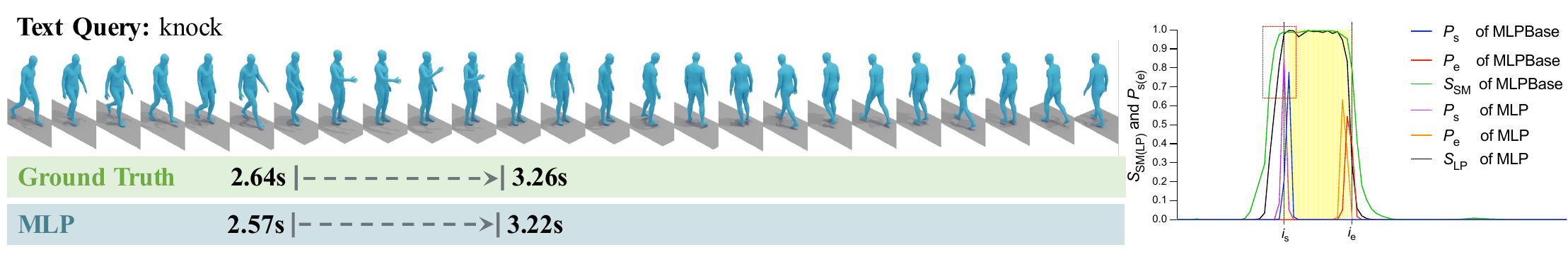}
\includegraphics[width=0.99\linewidth,trim=0 5 0 5,clip]{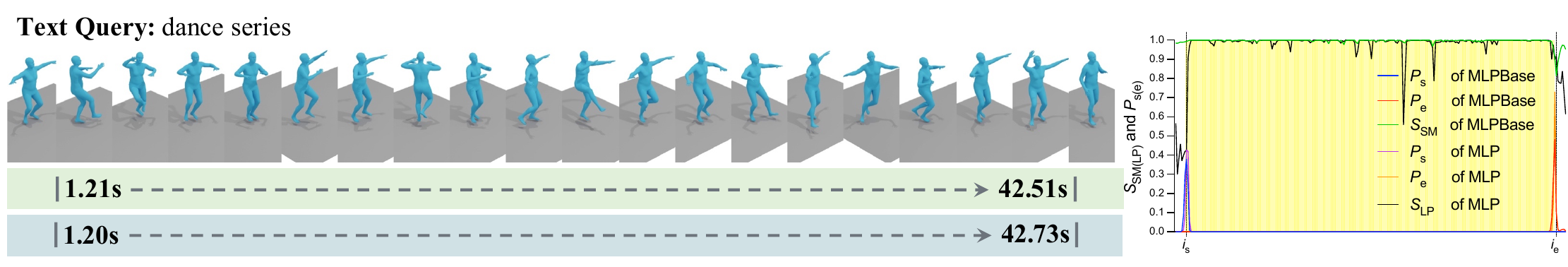}
\vspace{-2pt}
\caption{Qualitative grounding results on BABEL test set. We visualize the start/end probabilities $\mathcal{P}_{\text{s/e}}$ (Eq. \ref{eq:infer}) of \tesla{} and \teslam{}, and the highlight score $\mathcal{S}_{\text{LP/SM}}$ (Sec. \ref{subsubsec:lpmatcher}), on the right. It can be observed that in the foreground region (yellow background), the $\mathcal{S}_{\text{LP}}$ of \teslam{} is generally higher than the $\mathcal{S}_{\text{SM}}$ of \tesla{}.}
\label{fig:qualitative_success2}
\vspace{-0.3cm}
\end{figure*}

We start by describing the dataset and evaluation protocol used in our experiments (Sec.~\ref{subsec:db_ep}). We then report our implementation details (Sec.~\ref{subsec:implement_details}), and the performance of our model on TSLM benchmark (Sec.~\ref{subsec:benchmark}). Next, we introduce our ablation studies and in-depth analysis (Sec.~\ref{subsec:ablation}). Finally, we provide qualitative results for localization (Sec.~\ref{subsec:qualitative}).

\subsection{Dataset and Evaluation Protocol} \label{subsec:db_ep}

\noindent\textbf{BABEL}~\cite{punnakkal2021babel} provides frame-level natural language descriptions for the AMASS motion capture collection~\cite{mahmood2019amass}. We utilize all ``raw\_label" descriptions (\ie, text query), including extra files provided officially; however, we exclude query$=$``\textit{transition}" as it does not convey consistent semantic information for describing motion. The dataset is split into 60\% training data, 20\% validation data, and 20\% test data. Nevertheless, the test set is not publicly accessible as it is reserved for challenges. Therefore, we utilize 20\% of the validation set as the benchmark test set.

\noindent\textbf{HumanML3D (Restore)} is a dataset we adapt original HumanML3D (dubbed H3D) \cite{guo2022generating} for TSLM. The original H3D consists of segments extracted from the \textit{source motions} of AMASS, along with a corresponding description. We consider these segments as target moments for TSLM, with the source motions (before extraction) treated as new samples. This way, these segments are given temporal information. We filter out samples where the relative length of the target moment compared to the source motion exceeds 80\%. Data augmentation is then performed through left-right mirroring (both motion and corresponding queries). After these processes, HumanML3D (Restore) exhibits a 68\% difference from the source motions in BABEL. We found that a small portion of the restored H3D contains source motions that appear in both the training and testing sets. To prevent cheating, we re-split the dataset.

In both datasets, motions are processed at 20Hz. We extract joint positions using the SMPL layer~\cite{loper2015smpl} and apply the same pose processing steps as in~\cite{holden2017phase}. Their statistics are shown in Table. \ref{tab:datatset}.

\begin{figure*}[t]
\centering
\vspace{-2pt}
\includegraphics[width=0.99\linewidth,trim=0 5 0 5,clip]{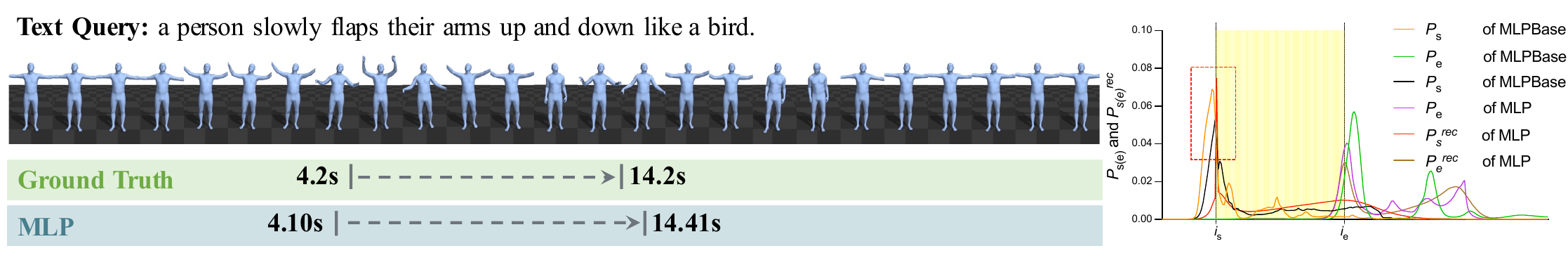}
\includegraphics[width=0.99\linewidth,trim=0 5 0 5,clip]{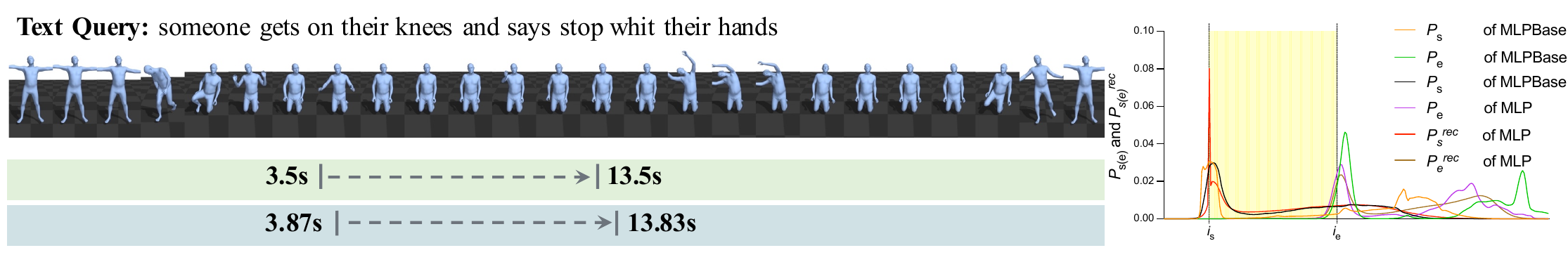}
\vspace{-2pt}
\caption{Qualitative grounding results on HumanML3D (Restore) test set. We visualize the probability distributions of the two parts in the LP-Predictor (Sec. \ref{subsubsec:lppredictor}). Please focus on the red rectangle on the right. It can be observed that, compared to $\mathcal{P}_{\text{s}}$ (\tesla{}), $\mathcal{P}_{\text{s}}$ (\teslam{}) is closer to $\mathcal{P}_{\text{s}}^{\text{rec}}$. This indicates that the predictions of \teslam{}, which incorporates prior knowledge, are more accurate.}
\vspace{-0.35cm}
\label{fig:qualitative_success1}
\end{figure*}

\noindent\textbf{Evaluation protocol.} {\ul We have defined two evaluation protocols}.
(a) \textbf{Normal}: We use ``IoU=$\mu$" and ``mIoU" as evaluation metrics. ``IoU=$\mu$" indicates the percentage of text queries with at least one retrieved moment having an intersection over union (IoU) greater than $\mu$ with the ground truth. ``mIoU" is the average IoU over all test samples. In our experiments, we use $\mu \in \{0.5, 0.7, 0.9\}$, and the retrieved moments are obtained by transforming the boundaries generated by the best located score (\ie, $\textrm{Recall}@1$) from Eq. \ref{eq:infer}.
(b) \textbf{Assigned}. Protocol (a) has limitations as it does not account for the possibility of repeated or similar actions within one motion. For instance, within one untrimmed motion, we may have A \{``a person walk forward", (1s, 3s)\} and B \{``a human moving forward", (9s, 15s)\}. When A is temporally located to (9s,15s), protocol (a) would result in a zero IoU, which is clearly unreasonable. To address this, we utilize an external large-scale language model to provide proxy similarity. Specifically, we use the popular MPNet\footnote{\url{https://huggingface.co/sentence-transformers/all-mpnet-base-v2}}~\cite{song2020mpnet} to compute the similarity between multiple text queries for each motion. Then, if the similarity is above a certain threshold, we consider its corresponding moment as a ``\textit{false-negative moment}". The final result is the maximum IoU between the retrieved moment and multiple false-negative moments.

\begin{table}[]
\centering
\small
\resizebox{\columnwidth}{!}{%
\begin{tabular}{lcccc}
\toprule
\multicolumn{5}{c}{\textbf{BABEL Dataset}} \\ 
\multicolumn{5}{c}{\textit{(a) Normal}} \\ \midrule
\multicolumn{1}{l|}{Method} & IoU@0.5 & IoU@0.7 & IoU@0.9 & mIoU \\ \midrule
\multicolumn{1}{l|}{$\text{TMR}^{\star}$~\cite{petrovich2023tmr}} & 15.96 &  7.55 &  1.92 & 17.70 \\
\multicolumn{1}{l|}{$\text{Rehamot}^{\star}$~\cite{yan2023cross}} & 14.49 &  5.75 &  1.10  & 17.19 \\
\multicolumn{1}{l|}{$\text{MMN}^{\dagger}$~\cite{wang2022negative}} & 28.86 & 20.86 & 7.95  & 27.10 \\ 
\multicolumn{1}{l|}{$\text{LGI}^{\dagger}$~\cite{mun2020local}} & 7.58 &  3.61 &  0.65  & 11.57 \\
\multicolumn{1}{l|}{$\text{VSLNet}^{\dagger}$~\cite{zhang2020span}} & 23.03  & 15.54 &  5.49  & 25.21 \\
\multicolumn{1}{l|}{$\text{SeqPAN}^{\dagger}$~\cite{zhang2021parallel}} & 27.84 &  20.58  & 7.45 &  28.66  \\ 
\multicolumn{1}{l|}{$\text{EAMAT}^{\dagger}$~\cite{yang2022entity}} & 27.34 &  19.60 &  7.63  &  27.86 \\ 
\multicolumn{1}{l|}{$\text{MS-DETR}^{\dagger}$~\cite{wang2023ms}} & {\ul 33.41} & 23.19 & 7.57  & 30.18 \\  \midrule
\multicolumn{1}{l|}{\textbf{MLPBase}} & 28.52  & 21.20 & 8.90  &  28.51 \\
\multicolumn{1}{l|}{\textbf{MLP-S}} & 32.08 &  24.67  & 10.45 &  30.50 \\  \midrule
\multicolumn{1}{l|}{\textbf{MLP}} & 33.22  & {\ul 25.79}  & {\ul 11.39} & {\ul 31.19} \\ 
\multicolumn{1}{l|}{\textbf{MLP} \fire{}} & \textbf{35.09} &  \textbf{27.18} &  \textbf{11.88} &  \textbf{32.63} \\ \midrule

\multicolumn{5}{c}{\textit{(b) Assigned}} \\ \midrule
\multicolumn{1}{l|}{$\text{TMR}^{\star}$~\cite{petrovich2023tmr}} & 29.03 &  13.63 &  3.41  & 31.49 \\
\multicolumn{1}{l|}{$\text{Rehamot}^{\star}$~\cite{yan2023cross}} & 26.18 &  10.51 &  2.07  & 30.41 \\
\multicolumn{1}{l|}{$\text{MMN}^{\dagger}$~\cite{wang2022negative}} & 46.30 & 34.39 & 13.61  & 43.84 \\ 
\multicolumn{1}{l|}{$\text{LGI}^{\dagger}$~\cite{mun2020local}} & 9.24 &  4.16 &  0.74  & 15.52 \\
\multicolumn{1}{l|}{$\text{VSLNet}^{\dagger}$~\cite{zhang2020span}} & 31.20 &  19.89  &6.64   & 34.50 \\
\multicolumn{1}{l|}{$\text{SeqPAN}^{\dagger}$~\cite{zhang2021parallel}} &39.71  & 28.80 &10.63   & 41.20 \\ 
\multicolumn{1}{l|}{$\text{EAMAT}^{\dagger}$~\cite{yang2022entity}} & 40.34 & 29.13 & 11.48  & 41.19 \\ 
\multicolumn{1}{l|}{$\text{MS-DETR}^{\dagger}$~\cite{wang2023ms}} & {\ul 53.50} & 36.78 & 12.14  & 48.85 \\  \midrule
\multicolumn{1}{l|}{\textbf{MLPBase}} & 42.42  & 31.76 &  13.96 &  42.97 \\
\multicolumn{1}{l|}{\textbf{MLP-S}} & 51.12  & 39.56 &  17.10 &  48.57 \\  \midrule
\multicolumn{1}{l|}{\textbf{MLP}} & 52.42 & {\ul 40.95}& {\ul 18.81} & {\ul 49.75} \\
\multicolumn{1}{l|}{\textbf{MLP} \fire{}} & \textbf{57.21} &  \textbf{44.13} & \textbf{19.65}  & \textbf{53.14}\\ \bottomrule
\end{tabular}
}
\vspace{-5pt}
\caption{\textbf{TSLM on the BABEL dataset.} $\star$: Evaluate using the temporal pyramid approach. $\dagger$: Re-implemented on the BABEL datset \fire{}: Fine-tune RoBERTa using LoRA technology to further improve performance.}
\label{tab:benchmark_babel}
\vspace{-20pt}
\end{table}

\begin{table}[]
\centering
\small
\resizebox{\columnwidth}{!}{%
\begin{tabular}{lcccc}
\toprule
\multicolumn{5}{c}{\textbf{HumanML3D Dataset (Restore)}} \\ 
\multicolumn{5}{c}{\textit{(a) Normal}} \\ \midrule
\multicolumn{1}{l|}{Method} & IoU@0.5 & IoU@0.7 & IoU@0.9 & mIoU \\ \midrule
\multicolumn{1}{l|}{$\text{TMR}^{\diamond}$~\cite{petrovich2023tmr}} & 44.01 &  29.03 & 5.29 & 40.40 \\
\multicolumn{1}{l|}{$\text{Rehamot}^{\diamond}$~\cite{yan2023cross}} & 42.28 &  28.64 & 4.60 & 39.87 \\
\multicolumn{1}{l|}{$\text{MMN}^{\dagger}$~\cite{wang2022negative}} & 46.84 &  34.70 & 9.62  & 43.75 \\ 
\multicolumn{1}{l|}{$\text{LGI}^{\dagger}$~\cite{mun2020local}} & 38.61 &  22.50 &  6.63  & 38.90 \\
\multicolumn{1}{l|}{$\text{VSLNet}^{\dagger}$~\cite{zhang2020span}} & 40.73 &  20.84 & 8.77 & 43.05 \\
\multicolumn{1}{l|}{$\text{SeqPAN}^{\dagger}$~\cite{zhang2021parallel}} & 44.15  & 27.84 &  10.78 & 45.20 \\
\multicolumn{1}{l|}{$\text{EAMAT}^{\dagger}$~\cite{yang2022entity}} & 45.56 & 33.02 & 10.55 & 44.80 \\
\multicolumn{1}{l|}{$\text{MS-DETR}^{\dagger}$~\cite{wang2023ms}} &  46.07 &  35.90 & 11.52  & 44.84 \\ \midrule
\multicolumn{1}{l|}{\textbf{MLPBase}} & 45.96 &  29.12 & 12.16 & 45.34 \\
\multicolumn{1}{l|}{\textbf{MLP-S}} & 46.04 &  35.85 & 14.10 & 45.05 \\  \midrule
\multicolumn{1}{l|}{\textbf{MLP}} & {\ul 46.88} & {\ul 41.14}  & {\ul 18.38} & {\ul 46.23} \\
\multicolumn{1}{l|}{\textbf{MLP} \fire{}} & \textbf{47.65} &  \textbf{42.56} & \textbf{19.07} & \textbf{46.93} \\ \midrule

\multicolumn{5}{c}{\textit{(b) Assigned}} \\ \midrule
\multicolumn{1}{l|}{$\text{TMR}^{\diamond}$~\cite{petrovich2023tmr}} & 62.25 &  41.85 & 7.17 & 53.60 \\
\multicolumn{1}{l|}{$\text{Rehamot}^{\diamond}$~\cite{yan2023cross}} & 58.73 &  38.81 & 5.63 & 52.16 \\
\multicolumn{1}{l|}{$\text{MMN}^{\dagger}$~\cite{wang2022negative}} &  74.88 &  57.29 &  15.35  & 63.94 \\ 
\multicolumn{1}{l|}{$\text{LGI}^{\dagger}$~\cite{mun2020local}} & 61.66 &  38.46 &  10.53  & 54.77 \\
\multicolumn{1}{l|}{$\text{VSLNet}^{\dagger}$~\cite{zhang2020span}} & 57.62 &  34.10 & 13.61 & 55.23 \\
\multicolumn{1}{l|}{$\text{SeqPAN}^{\dagger}$~\cite{zhang2021parallel}} & 62.76  & 42.53 & 16.22  & 58.71 \\
\multicolumn{1}{l|}{$\text{EAMAT}^{\dagger}$~\cite{yang2022entity}} & 64.79 & 50.81  & 15.82 & 59.70 \\
\multicolumn{1}{l|}{$\text{MS-DETR}^{\dagger}$~\cite{wang2023ms}} & 73.00 &  57.94 & 17.87  & 64.02 \\ \midrule
\multicolumn{1}{l|}{\textbf{MLPBase}} & 66.52 &  47.32 & 19.48 & 60.74 \\
\multicolumn{1}{l|}{\textbf{MLP-S}} & 74.45 &  60.67 & 23.09 & 65.64 \\  \midrule
\multicolumn{1}{l|}{\textbf{MLP}} & {\ul 75.03} &  {\ul 68.91} & {\ul 29.05} & {\ul 68.34} \\
\multicolumn{1}{l|}{\textbf{MLP} \fire{}} & \textbf{76.94} &  \textbf{71.17} & \textbf{31.54} & \textbf{69.55} \\ \bottomrule
\end{tabular}
}
\vspace{-5pt}
\caption{\textbf{TSLM on the HumanML3D (Restore) dataset.} $\diamond$:  The original retrieval models were trained using contrastive learning on the HumanML3D dataset (the original dataset, not our restored dataset). Following the original (official) split, cheating scenarios would arise in TSLM evaluations. Therefore, we retrained their models according to the split of our restored dataset and evaluated them using the temporal pyramid method. $\dagger$: Re-implemented on the HumanML3D (Restore) dataset \fire{}: Fine-tune RoBERTa using LoRA technology to further improve performance.}
\label{tab:benchmark_h3d}
\vspace{-0.5cm}
\end{table}

\subsection{Implementation Details} \label{subsec:implement_details}

We employ the AdamW optimizer~\cite{loshchilov2017decoupled} to guide the optimization process with parameters $[\beta_{1},\beta_{2}]=[0.9,0.999]$, a batch size of 64, and a weight decay parameter of 0.01. The model training process spans 100 epochs, with an initial learning rate of $2e-4$ and decayed by a factor of 10 at the 51th epoch. We utilize the frozen pre-trained model RoBERTa~\cite{liu2019roberta} to encode text queries into a series of  $d_{w}=768$ word embeddings and fine-tune it using LoRA technology~\cite{hu2021lora}. We set the number of interval snippets $S$ to 256, the hidden dimension $d$ to 256, the number of SGPA blocks $N_{\text{SGPA}}$ to 4, and include $L=7$ graph convolution layers in S-Enc. The perturb-rate $\alpha$ is set to 0.8. The flipping rate $\beta$ for BABEL and HumanML3D (Restore) was set to 0.1 and 0.2, respectively. The threshold for the protocol (b) Assigned in both datasets is empirically set to 0.995 and 0.8, respectively. All experiments are conducted on a single NVIDIA RTX 4090.

\begin{figure*}[!t]
\centering
\includegraphics[width=0.99\linewidth]{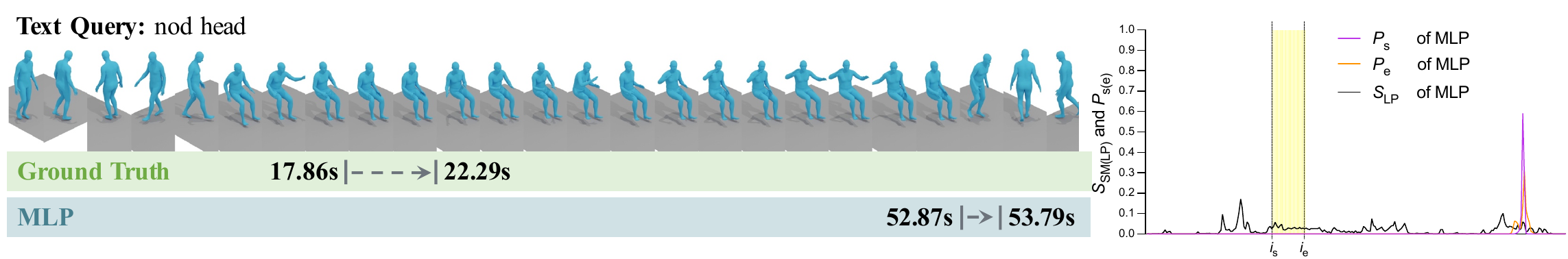}
\includegraphics[width=0.99\linewidth]{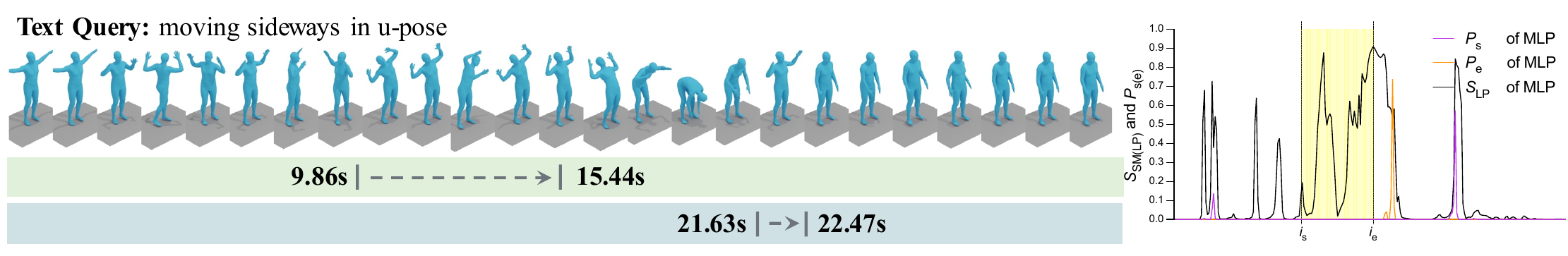}
\vspace{-2pt}
\caption{Qualitative grounding results on BABEL: Two failure cases. For example, in the second case, \teslam{} fails to grasp the meaning of ``moving sideways" deeply, as it only identifies the ``u-pose" posture without any actual movement. For further analysis, please refer to Sec. \ref{subsec:qualitative}.}
\label{fig:qualitative_failure}
\vspace{-15pt}
\end{figure*}

\begin{table}[]
\centering
\small
\resizebox{\columnwidth}{!}{%
\begin{tabular}{@{}cccccc@{}}
\toprule
\multicolumn{2}{c|}{\textbf{Components}} & \multirow{2}{*}{IoU@0.5} & \multirow{2}{*}{IoU@0.7} & \multirow{2}{*}{mIoU} & \multirow{2}{*}{} \\
\multicolumn{1}{c|}{LP-Matcher} & \multicolumn{1}{c|}{LP-Predictor} &  &  &  &  \\ \midrule
\xmark & \multicolumn{1}{c|}{\xmark} & 42.42 & 31.76 & 42.97 & $\diamondsuit$ \\
\xmark & \multicolumn{1}{c|}{\cmark} & 44.94 & 33.68 & 44.63 &  \\
\cmark & \multicolumn{1}{c|}{\xmark} & 49.06 & 36.77 & 47.22 &  \\
\cmark & \multicolumn{1}{c|}{\cmark} & \textbf{51.12} & \textbf{39.56} & \textbf{48.57} & $\heartsuit$ \\ \midrule
\multicolumn{6}{l}{\begin{tabular}[c]{@{}l@{}}$\diamondsuit$: \tesla{}  $\heartsuit$: \teslams{} \end{tabular}}
\end{tabular}%
}
\vspace{-5pt}
\caption{Ablation results for major contributors on BABEL, including LP-Matcher and LP-Predictor.}
\label{tab:ablation_main}
\vspace{-0.4cm}
\end{table}

\begin{figure}[ht]
\centering
\includegraphics[width=0.9\linewidth]{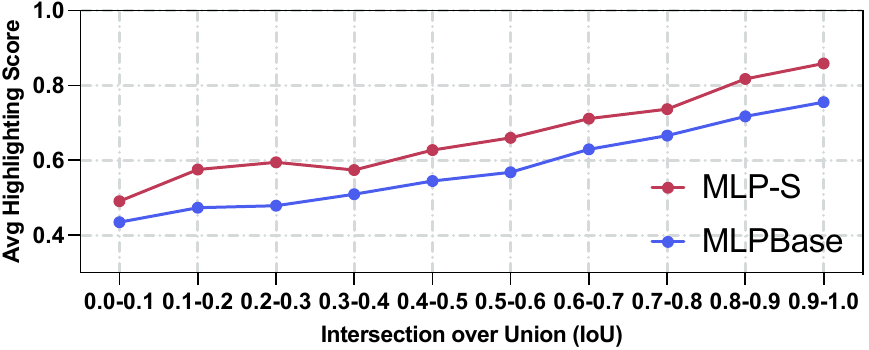}
\vspace{-5pt}
\caption{Line chart of the average highlighting scores of the foreground regions under different IoUs for \tesla{} and \teslam{}, on BABEL.}
\label{fig:avg_higghlight_score}
\vspace{-15pt}
\end{figure}

\subsection{A new benchmark \& Comparison to prior work} \label{subsec:benchmark}

\noindent\textbf{Baselines and MLP variants}. We present the performance of \teslam{} on the new localization benchmark, across two evaluation protocols. 
To obtain comparable baselines, we evaluate the recent retrieval models TMR \cite{petrovich2023tmr} and Rehamot \cite{yan2023cross} on the TSLM task using the temporal pyramid method~\cite{gao2017tall}, with a sliding window ranging from 20 to 200 frames and a stride of 5 frames. IoU is calculated by selecting the window size and position with maximum similarity to the text query.
Additionally, we compare against TSVL methods, including various representative works analyzed in the introduction: MMN~\cite{wang2022negative} (Anchor-based), LGI~\cite{mun2020local} (Regression-based), and VSLNet~\cite{zhang2020span}, SeqPAN~\cite{zhang2021parallel} (Span-based). Furthermore, we also include recent state-of-the-art methods: MS-DETR~\cite{wang2023ms} and EAMAT~\cite{yang2022entity}. We retrain them on the TSLM benchmark and report the best results.
{\ul To emphasize contributions, we design a generic baseline model named \tesla{}. Compared to \teslam{}, \tesla{} \textbf{excludes} the following three components:}
\textbf{A)} w/o S-Enc, directly uses an $FFN$ layer for motion-text dimension alignment.
\textbf{B)} w/o LP-Matcher: retaining the sequence matching approach of TSLV (Eq.~\ref{eq:highlighting}).
\textbf{C)} w/o LP-Predictor, retaining only the common Predicting part (Sec.~\ref{subsubsec:lppredictor}).
Additionally, as the other compared baselines do not provide spatial encoding for motion, a variant only removing \textbf{A)}, \teslams{}, is released for fair comparison.

\begin{table*}[htbp]
\centering
\resizebox{\textwidth}{!}{%
\begin{tabular}{m{0.15\textwidth}<{\raggedright}|>{\raggedright\arraybackslash}c|>{\raggedright\arraybackslash}c}
\toprule
\multicolumn{1}{c|}{\textbf{Text Query}} & \multicolumn{1}{c|}{\textbf{GT}} & \multicolumn{1}{c}{\textbf{Top-10 Retrieved Moments}} \\ \midrule
 {\Large \textit{climbing up}} & 
\raisebox{-.4\height}{\includegraphics[width=0.1\linewidth,trim=190 150 190 80,clip]{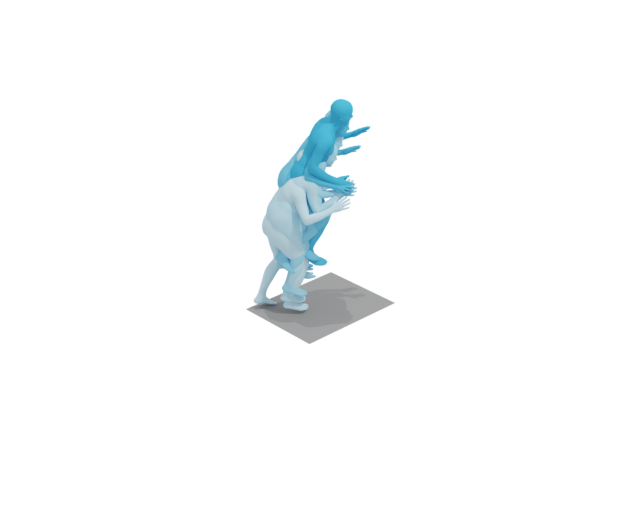}}
&
\raisebox{-.4\height}{
 \includegraphics[width=0.1\linewidth,trim=190 150 190 80,clip]{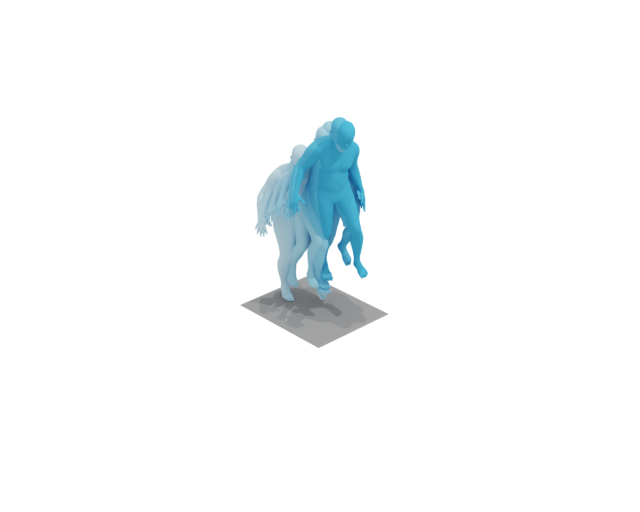}
\includegraphics[width=0.1\linewidth,trim=190 150 190 80,clip]{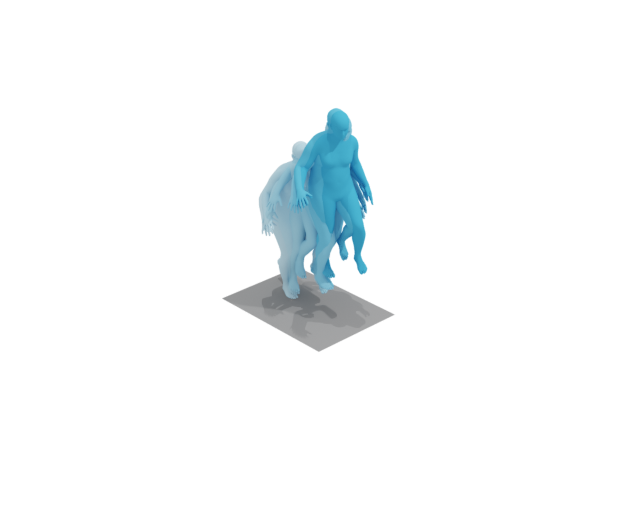}
\includegraphics[width=0.1\linewidth,trim=190 150 190 80,clip]{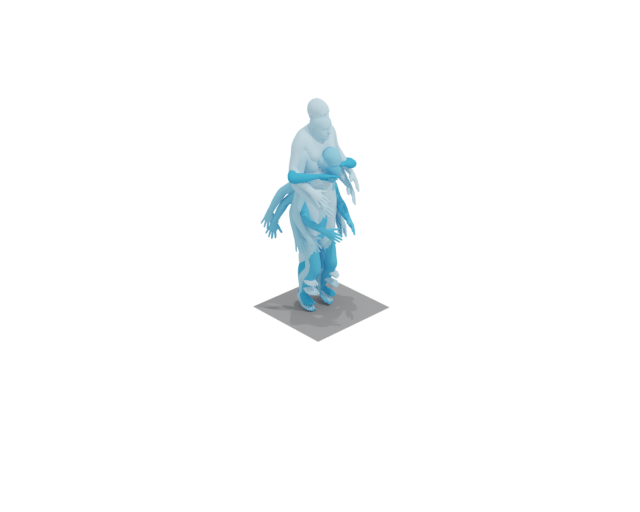}
\includegraphics[width=0.1\linewidth,trim=190 150 190 80,clip]{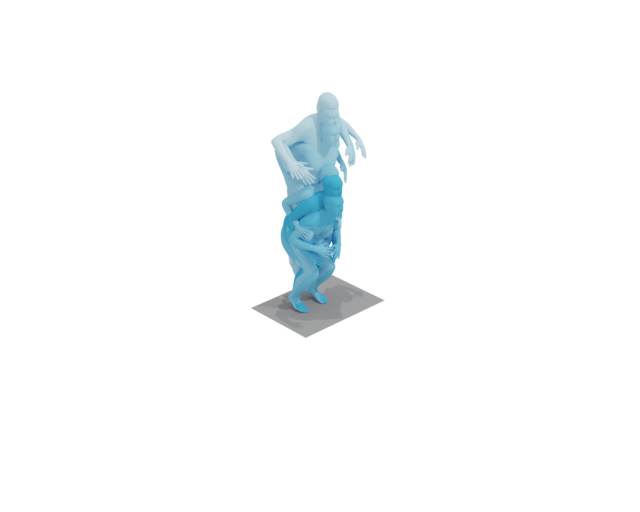}
\includegraphics[width=0.1\linewidth,trim=190 150 190 80,clip]{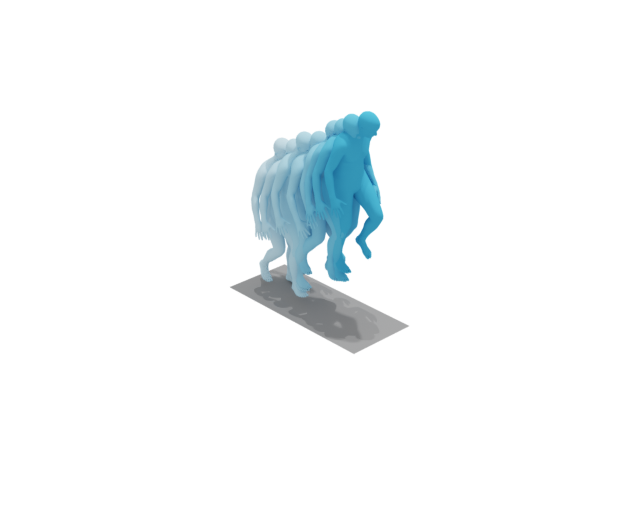}
\includegraphics[width=0.1\linewidth,trim=190 150 190 80,clip]{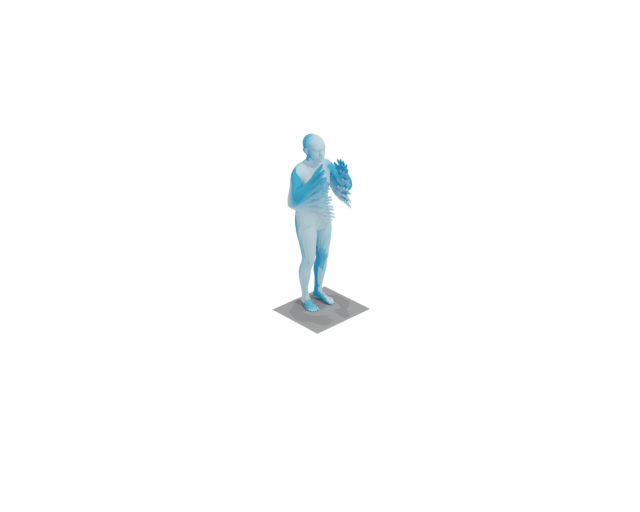}
\includegraphics[width=0.1\linewidth,trim=190 150 190 80,clip]{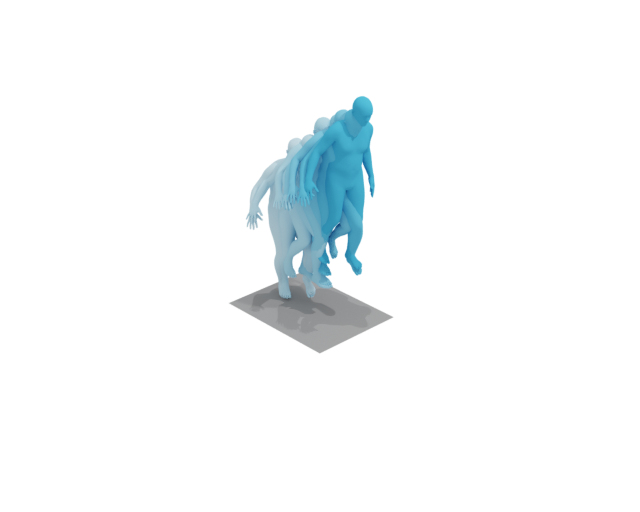}
\includegraphics[width=0.1\linewidth,trim=190 150 190 80,clip]{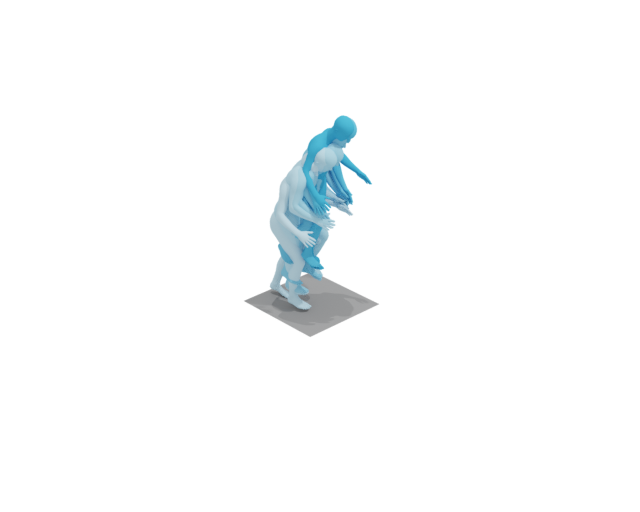}
\includegraphics[width=0.1\linewidth,trim=190 150 190 80,clip]{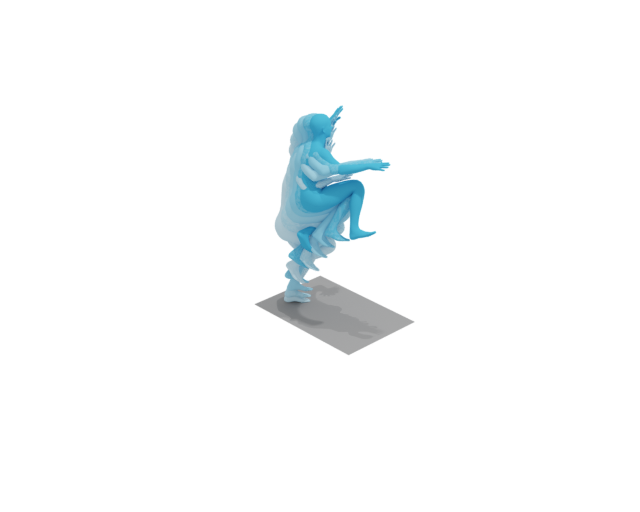}
\includegraphics[width=0.1\linewidth,trim=190 150 190 80,clip]{figs/use_case/7062_climbing_up/3906_mesh.png}} \vspace{2pt}
\\ \hline 
 {\Large \textit{jump kick right}} & 
\raisebox{-.4\height}{\includegraphics[width=0.1\linewidth,trim=190 150 190 80,clip]{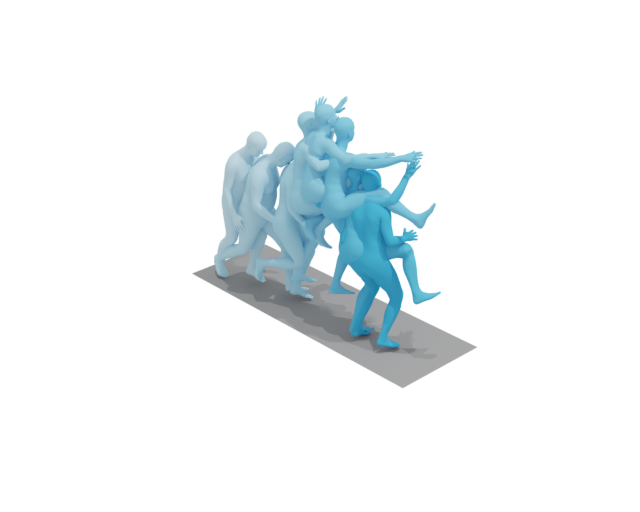}}
&
\raisebox{-.4\height}{
 \includegraphics[width=0.1\linewidth,trim=190 150 190 80,clip]{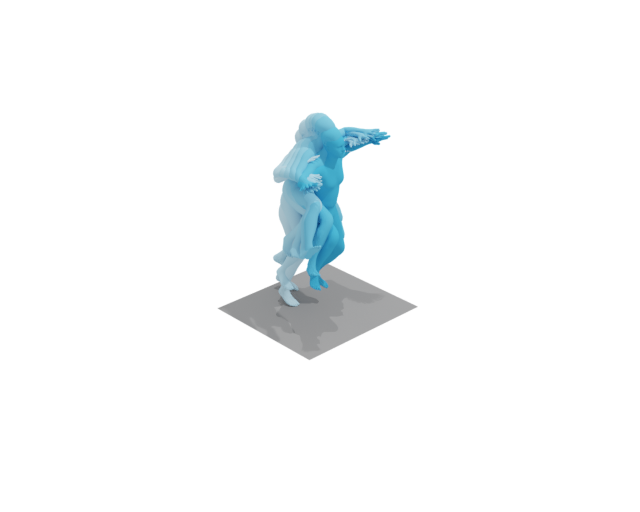}
\includegraphics[width=0.1\linewidth,trim=190 150 190 80,clip]{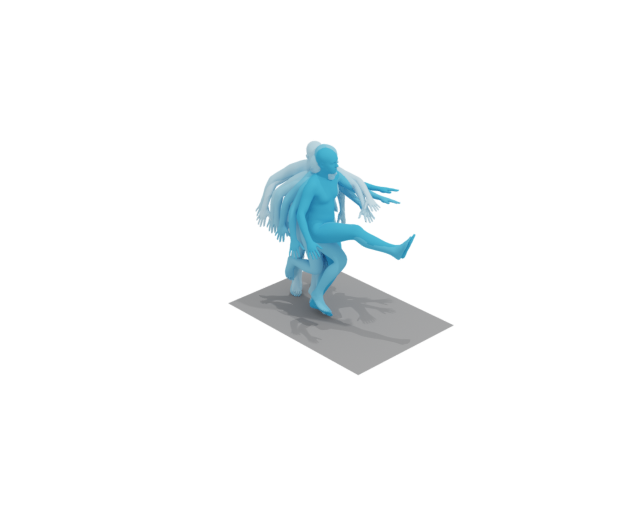}
\includegraphics[width=0.1\linewidth,trim=190 150 190 80,clip]{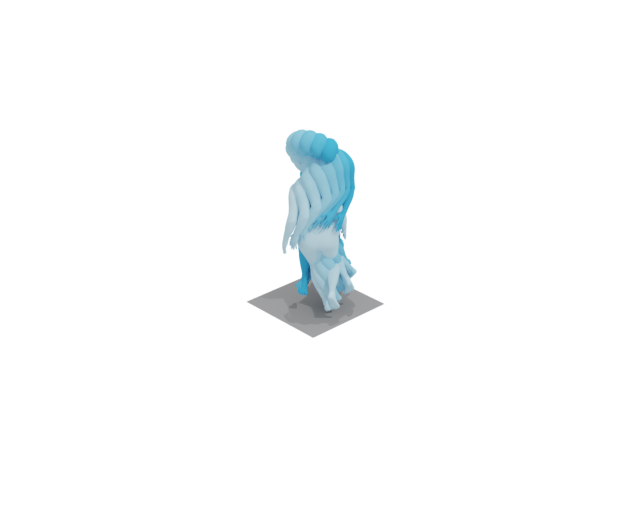}
\includegraphics[width=0.1\linewidth,trim=190 150 190 80,clip]{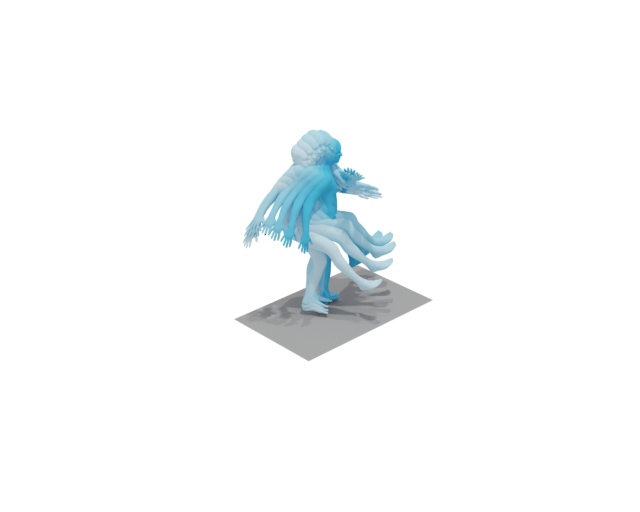}
\includegraphics[width=0.1\linewidth,trim=190 150 190 80,clip]{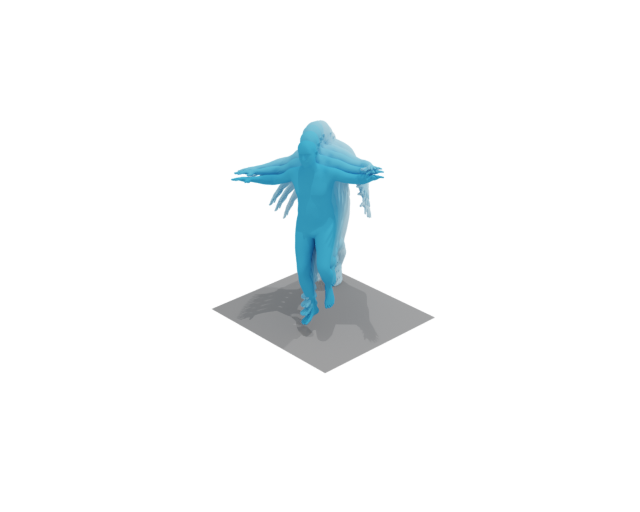}
\includegraphics[width=0.1\linewidth,trim=190 150 190 80,clip]{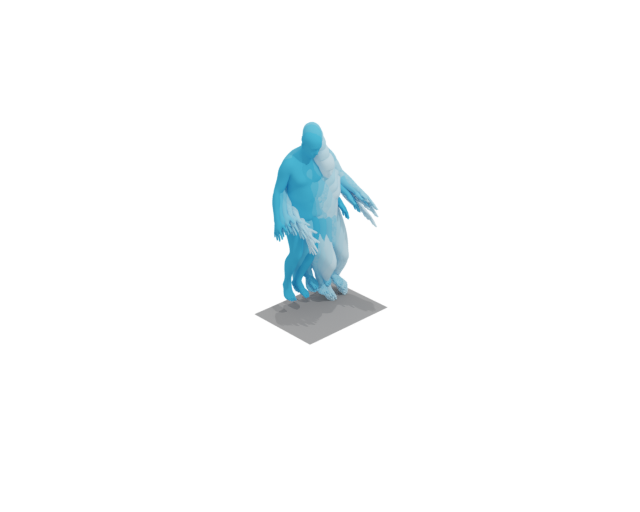}
\includegraphics[width=0.1\linewidth,trim=190 150 190 80,clip]{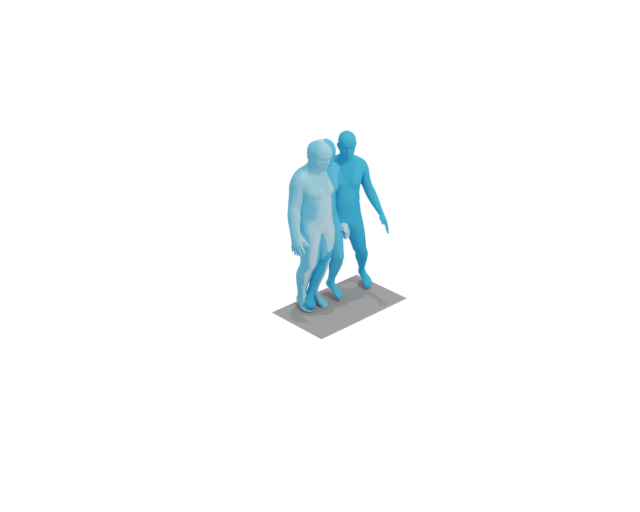}
\includegraphics[width=0.1\linewidth,trim=190 150 190 80,clip]{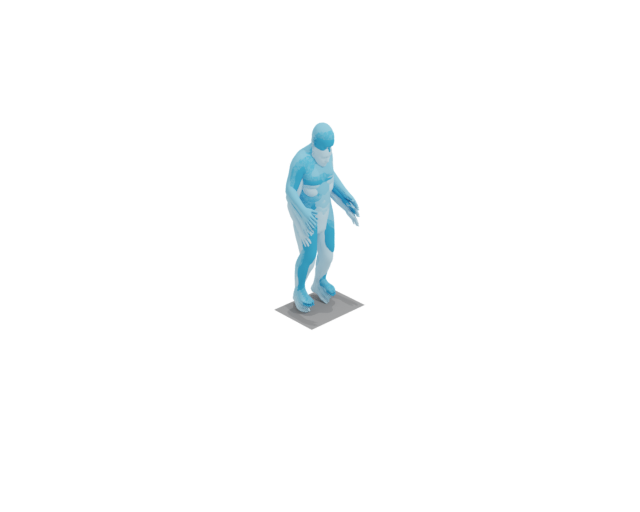}
\includegraphics[width=0.1\linewidth,trim=190 150 190 80,clip]{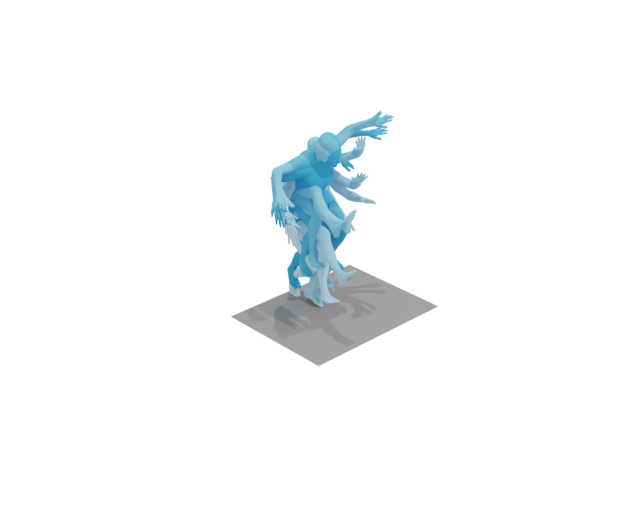}
\includegraphics[width=0.1\linewidth,trim=190 150 190 80,clip]{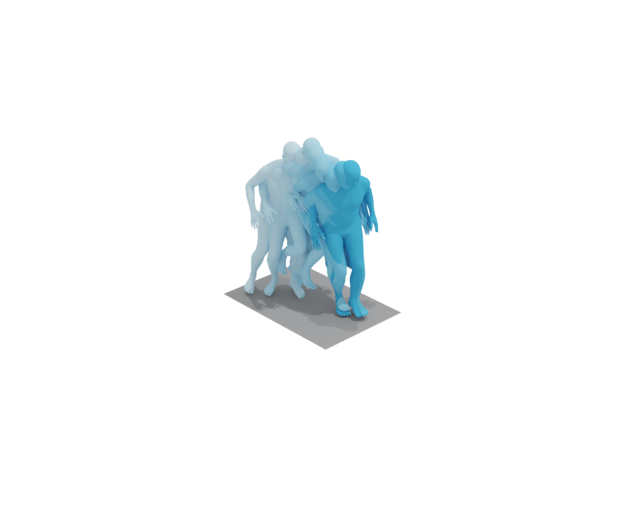}} \vspace{2pt}
\\ \hline 
{\Large \textit{practicing golf swing} } & 
\raisebox{-.4\height}{\includegraphics[width=0.1\linewidth,trim=190 150 190 80,clip]{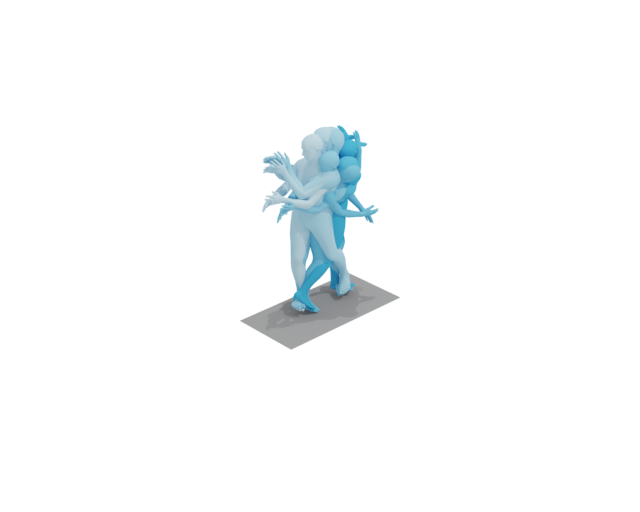}}
&
\raisebox{-.4\height}{
 \includegraphics[width=0.1\linewidth,trim=190 150 190 80,clip]{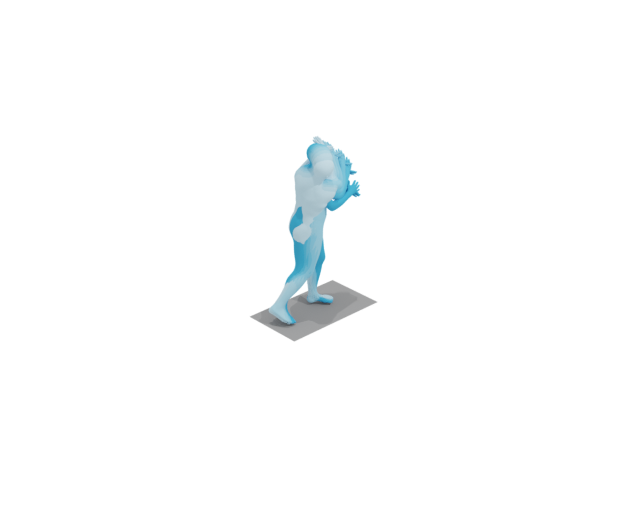}
\includegraphics[width=0.1\linewidth,trim=190 150 190 80,clip]{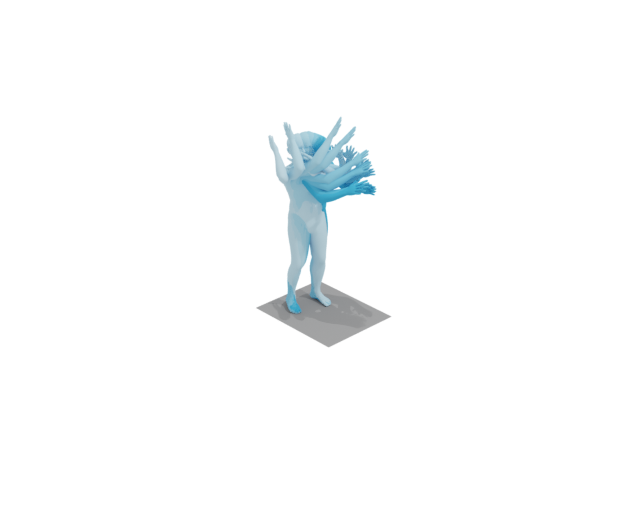}
\includegraphics[width=0.1\linewidth,trim=190 150 190 80,clip]{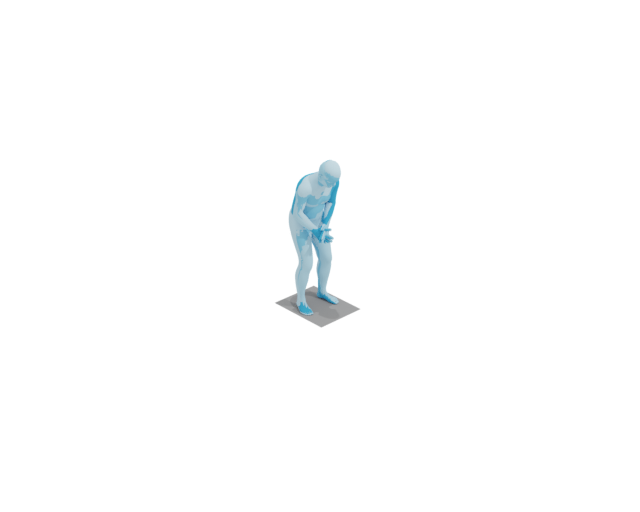}
\includegraphics[width=0.1\linewidth,trim=190 150 190 80,clip]{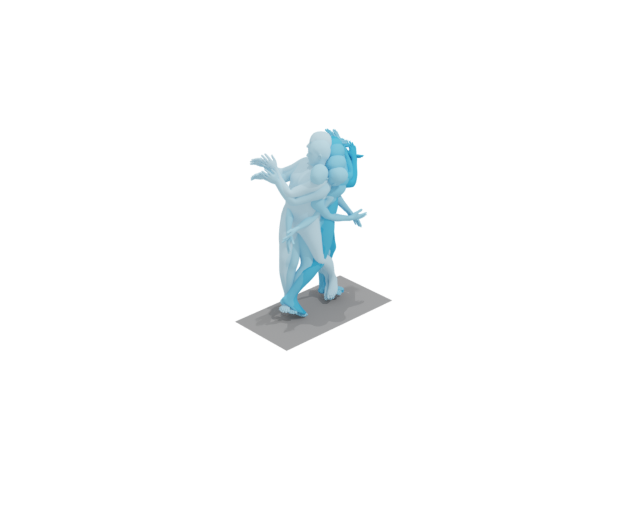}
\includegraphics[width=0.1\linewidth,trim=190 150 190 80,clip]{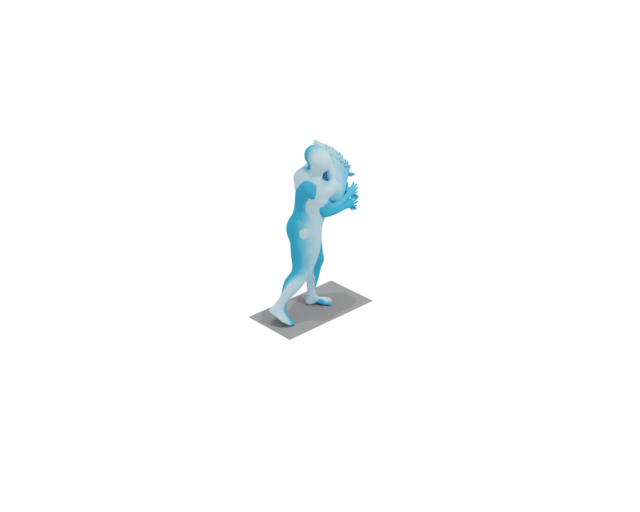}
\includegraphics[width=0.1\linewidth,trim=190 150 190 80,clip]{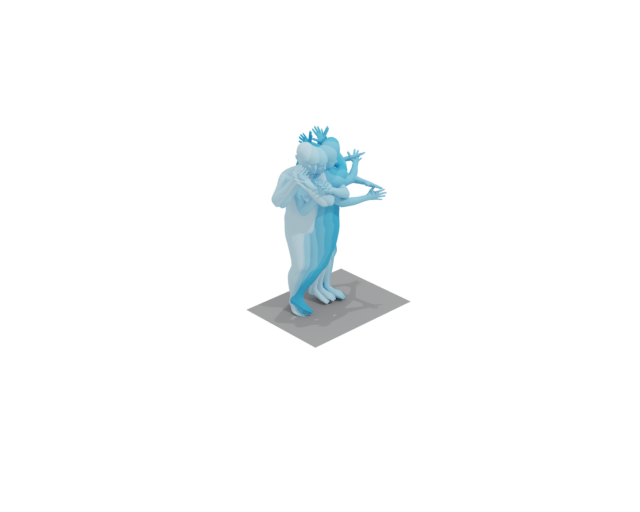}
\includegraphics[width=0.1\linewidth,trim=190 150 190 80,clip]{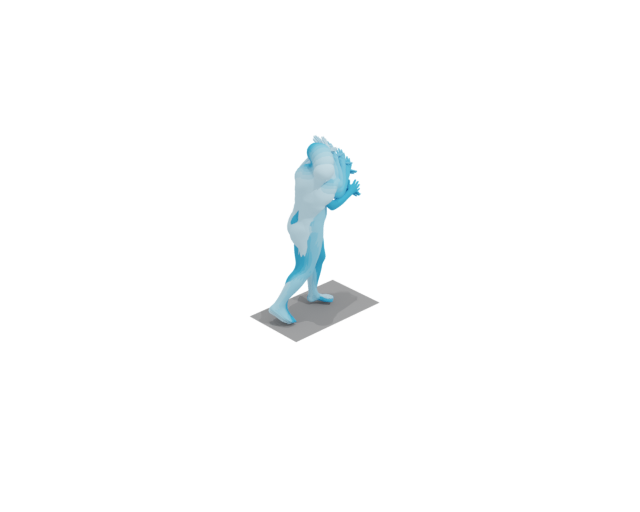}
\includegraphics[width=0.1\linewidth,trim=190 150 190 80,clip]{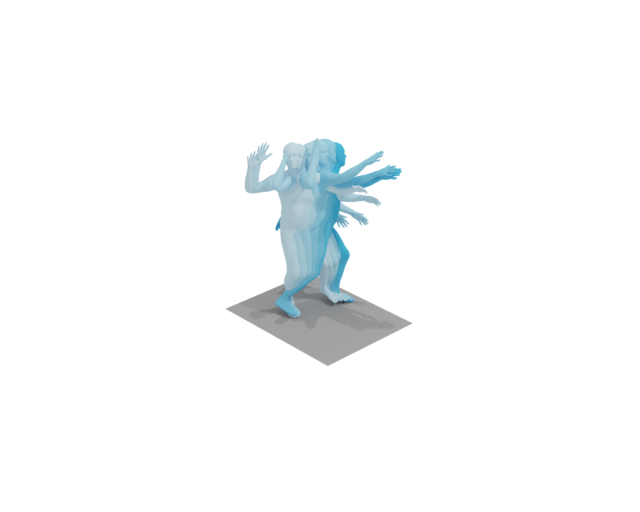}
\includegraphics[width=0.1\linewidth,trim=190 150 190 80,clip]{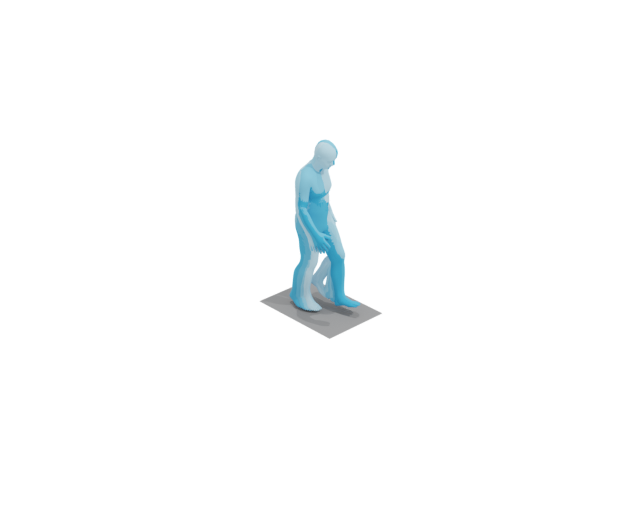}
\includegraphics[width=0.1\linewidth,trim=190 150 190 80,clip]{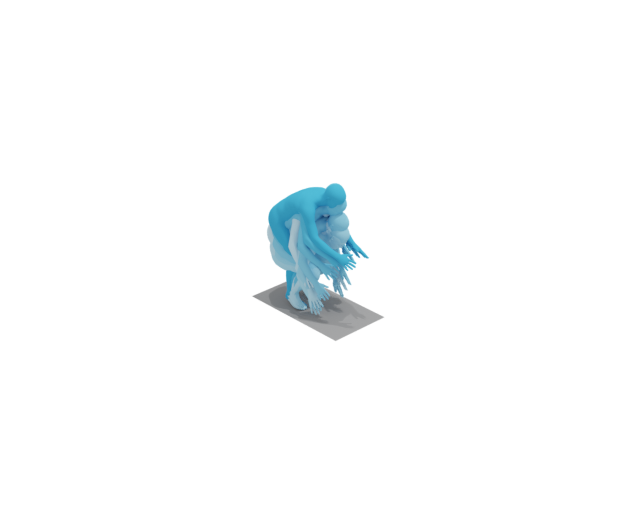}} \vspace{2pt}
\\ \hline 
{\Large \textit{walk in circles}} & 
\raisebox{-.4\height}{\includegraphics[width=0.1\linewidth,trim=120 100 120 80,clip]{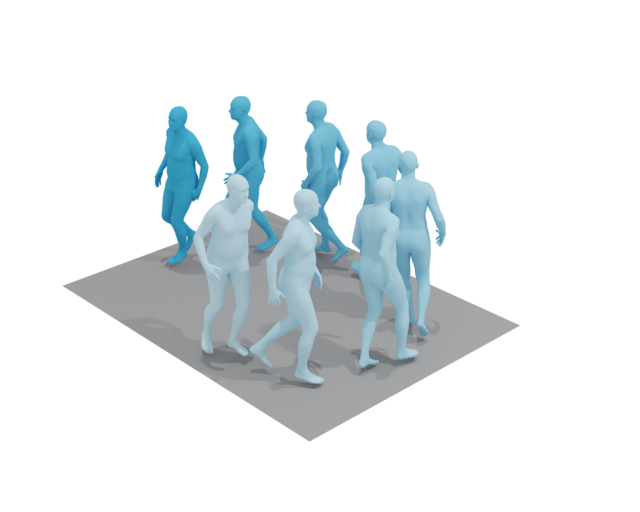}}
&
\raisebox{-.4\height}{
 \includegraphics[width=0.1\linewidth,trim=120 100 120 80,clip]{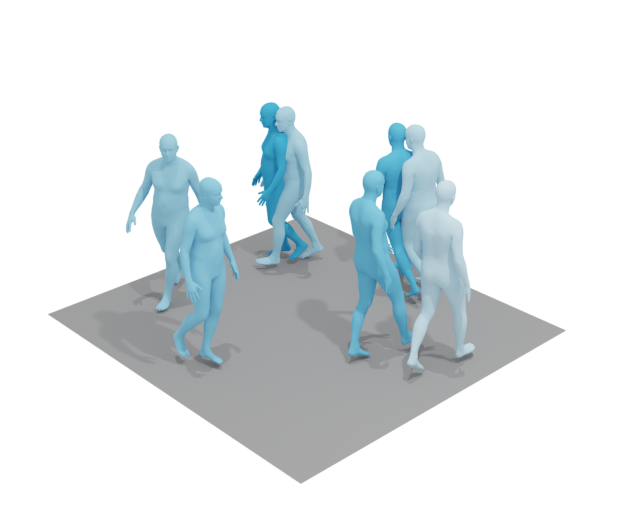}
\includegraphics[width=0.1\linewidth,trim=120 100 120 80,clip]{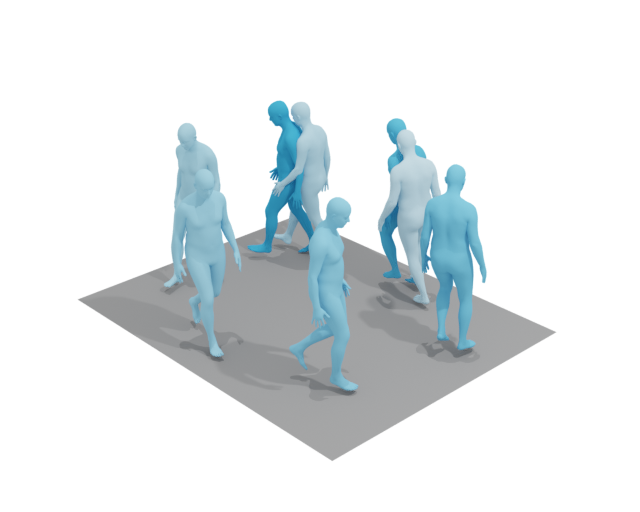}
\includegraphics[width=0.1\linewidth,trim=120 100 120 80,clip]{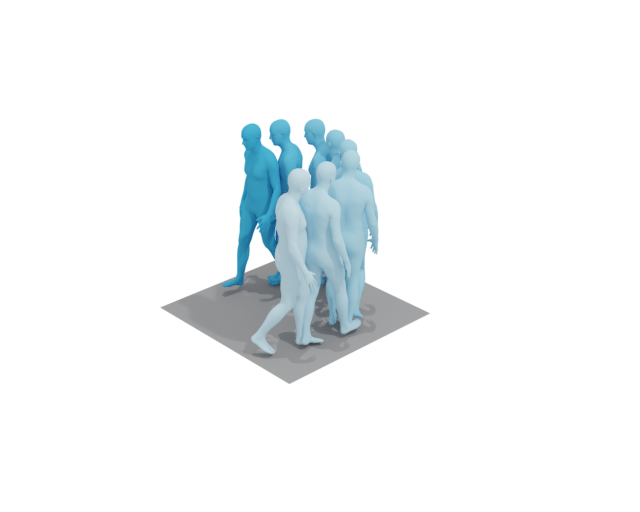}
\includegraphics[width=0.1\linewidth,trim=120 100 120 80,clip]{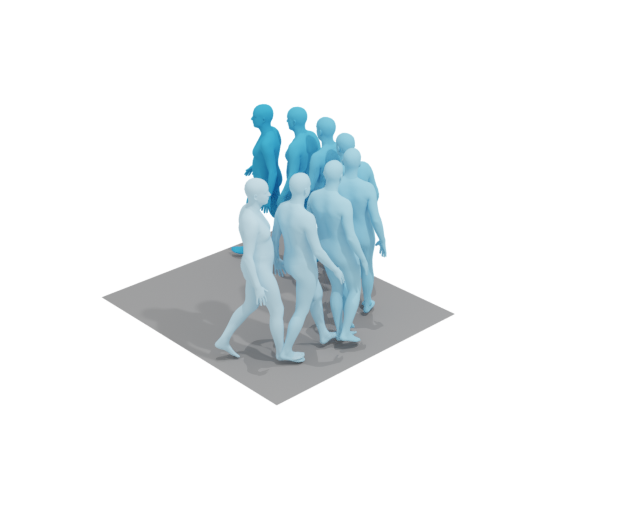}
\includegraphics[width=0.1\linewidth,trim=120 100 120 80,clip]{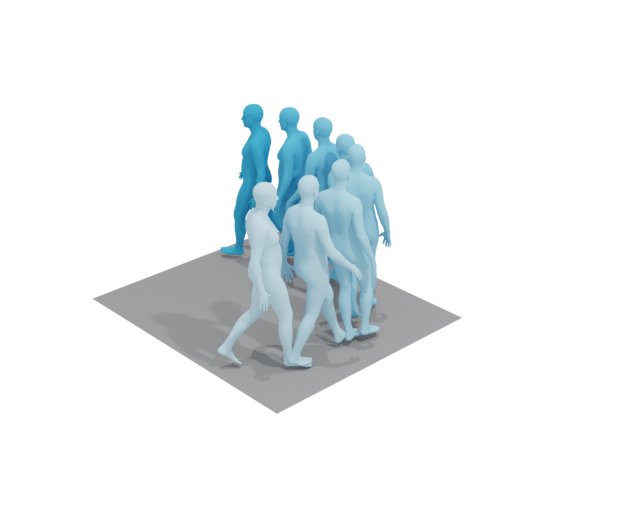}
\includegraphics[width=0.1\linewidth,trim=120 100 120 80,clip]{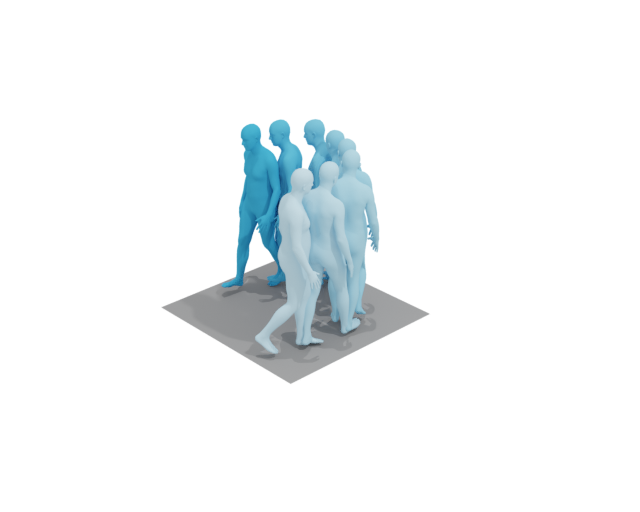}
\includegraphics[width=0.1\linewidth,trim=120 100 120 80,clip]{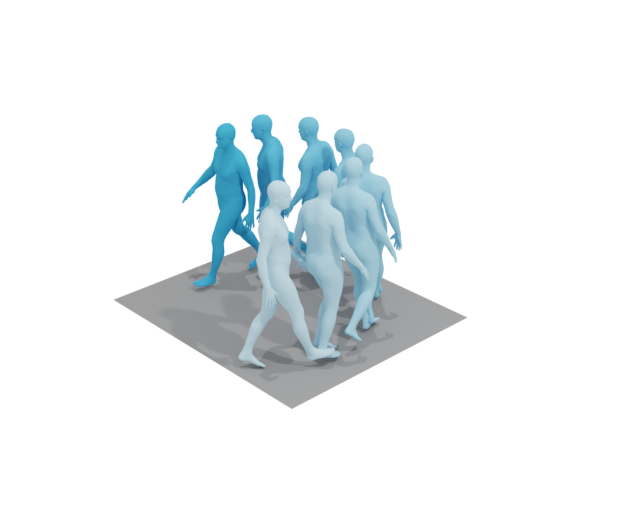}
\includegraphics[width=0.1\linewidth,trim=120 100 120 80,clip]{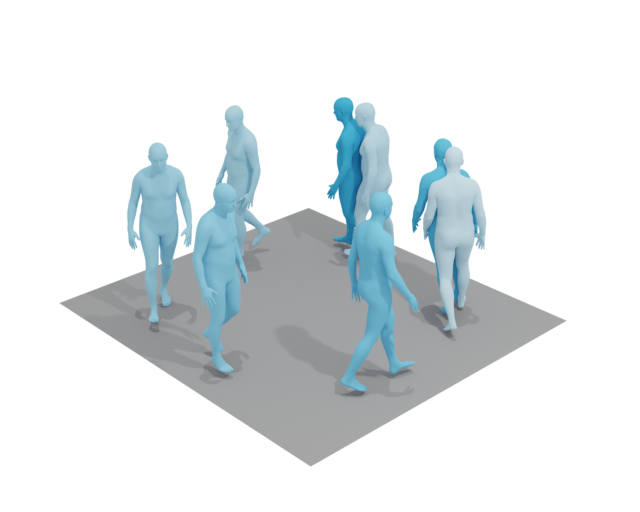}
\includegraphics[width=0.1\linewidth,trim=120 100 120 80,clip]{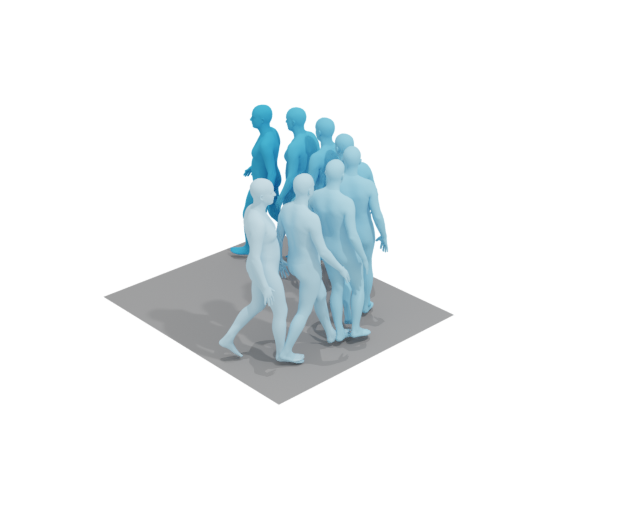}
\includegraphics[width=0.1\linewidth,trim=120 100 120 80,clip]{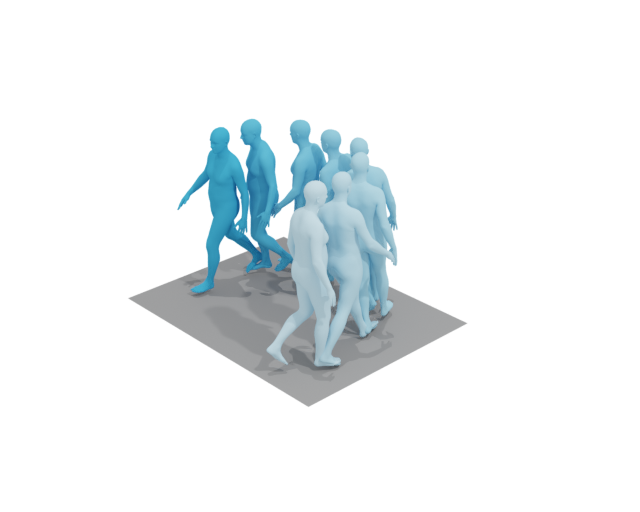}} \vspace{2pt}
\\ \hline 
{\Large \textit{standing on the left leg}} & 
\raisebox{-.4\height}{\includegraphics[width=0.1\linewidth,trim=190 150 190 80,clip]{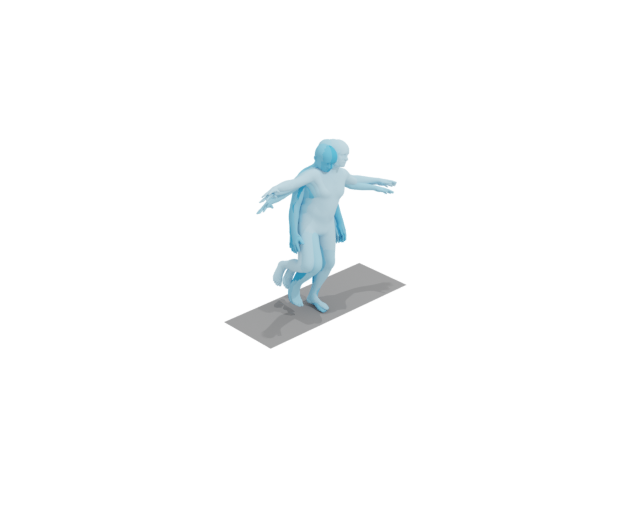}}
&
\raisebox{-.4\height}{
 \includegraphics[width=0.1\linewidth,trim=190 150 190 80,clip]{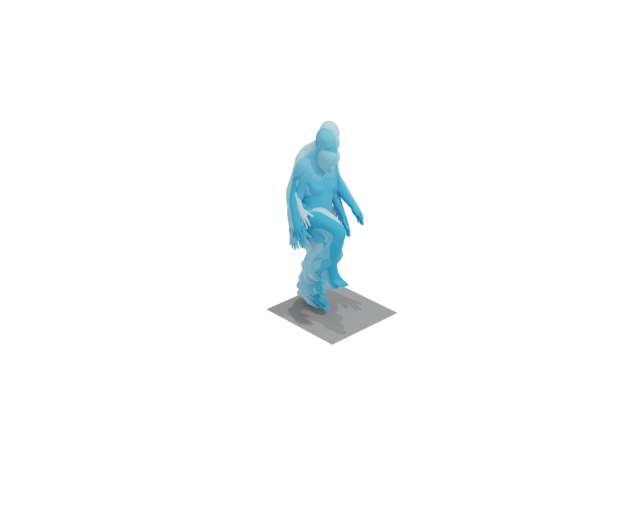}
\includegraphics[width=0.1\linewidth,trim=190 150 190 80,clip]{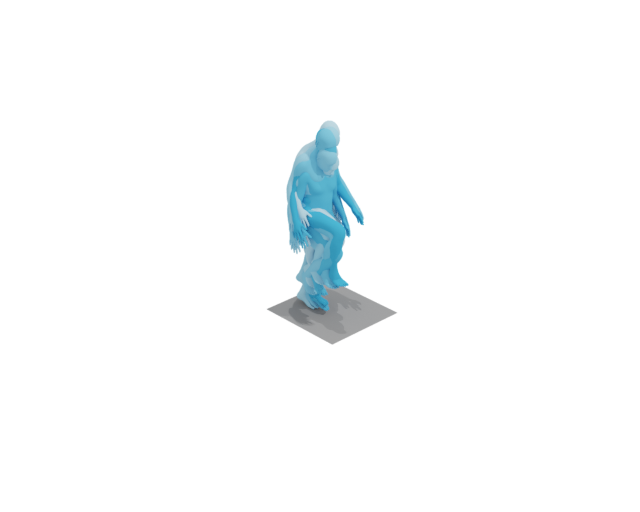}
\includegraphics[width=0.1\linewidth,trim=190 150 190 80,clip]{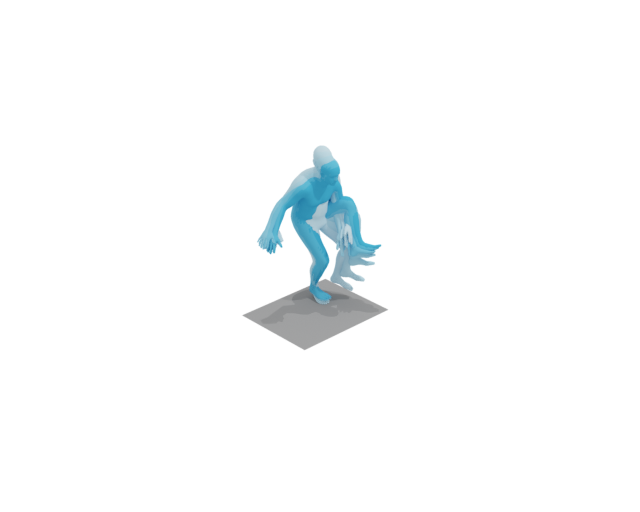}
\includegraphics[width=0.1\linewidth,trim=190 150 190 80,clip]{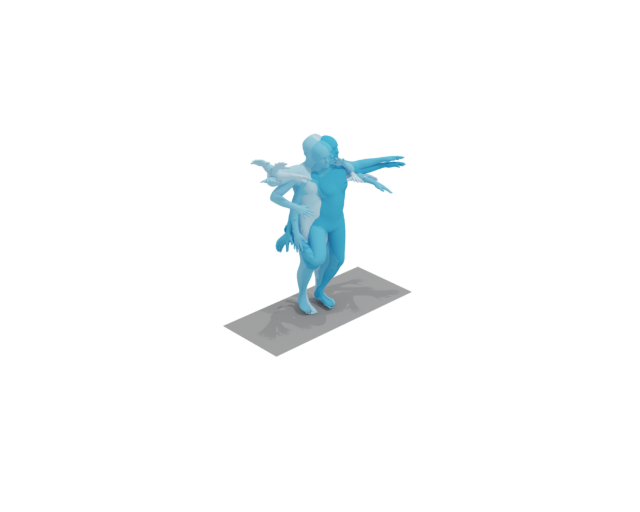}
\includegraphics[width=0.1\linewidth,trim=190 150 190 80,clip]{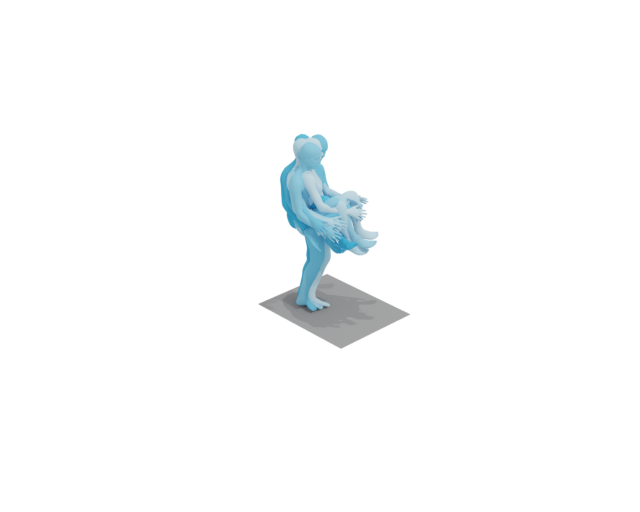}
\includegraphics[width=0.1\linewidth,trim=190 150 190 80,clip]{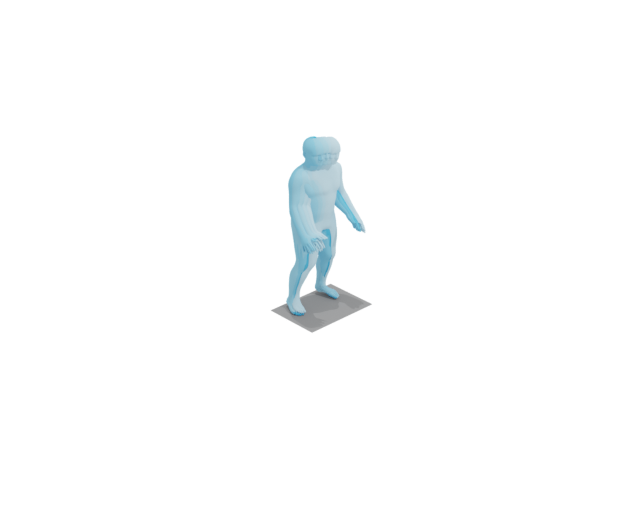}
\includegraphics[width=0.1\linewidth,trim=190 150 190 80,clip]{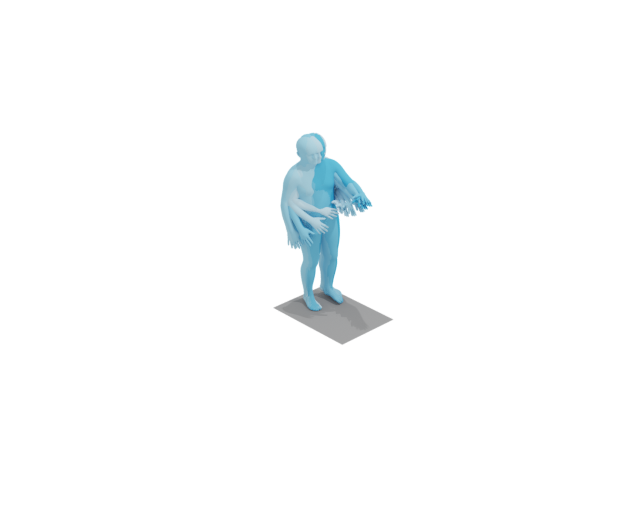}
\includegraphics[width=0.1\linewidth,trim=190 150 190 80,clip]{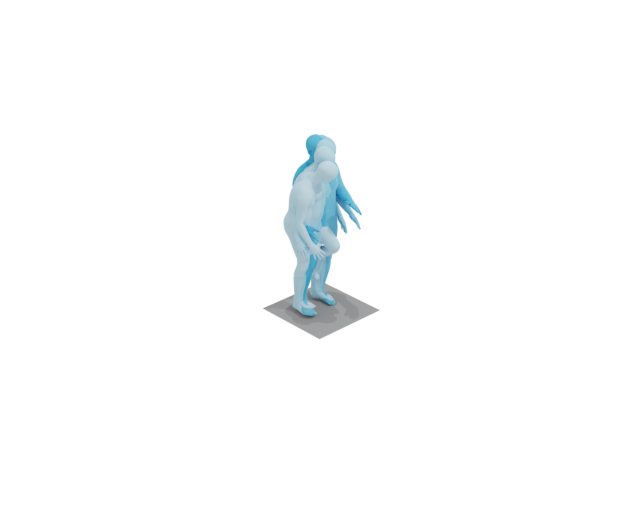}
\includegraphics[width=0.1\linewidth,trim=190 150 190 80,clip]{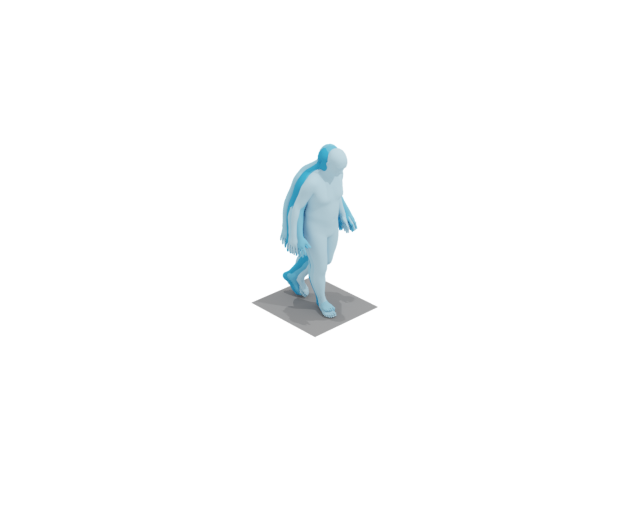}
\includegraphics[width=0.1\linewidth,trim=190 150 190 80,clip]{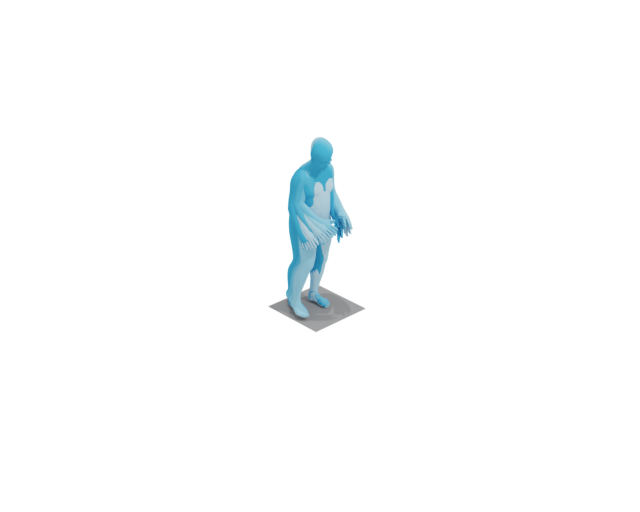}} \vspace{2pt}
\\ \hline
{\large \textit{standing with knees bent down with right arm extended}} & 
\raisebox{-.4\height}{\includegraphics[width=0.1\linewidth,trim=190 150 190 80,clip]{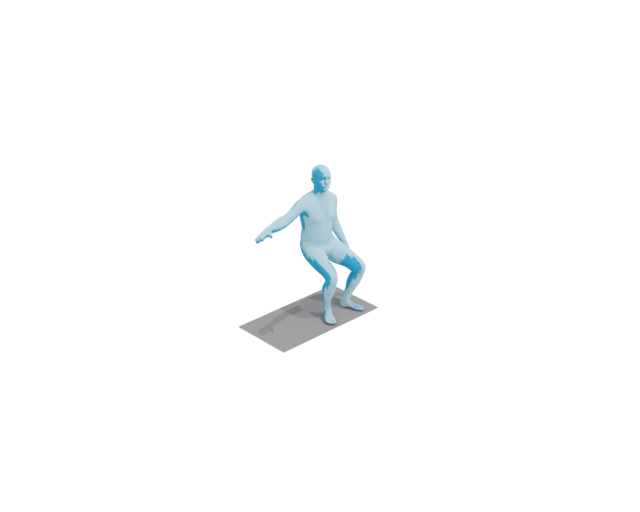}}
&
\raisebox{-.4\height}{
 \includegraphics[width=0.1\linewidth,trim=190 150 190 80,clip]{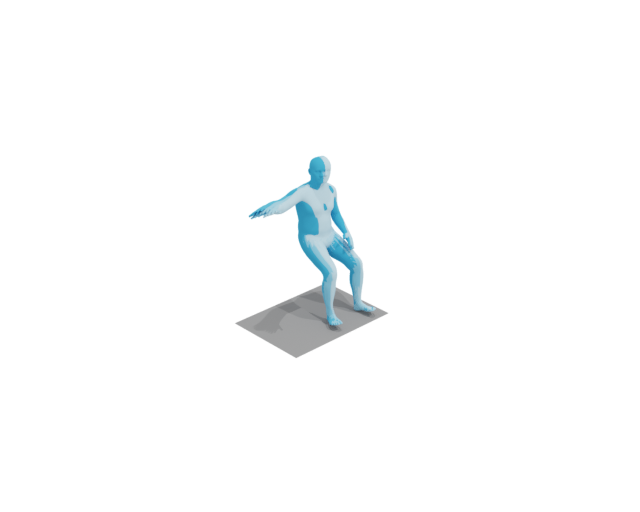}
\includegraphics[width=0.1\linewidth,trim=190 150 190 80,clip]{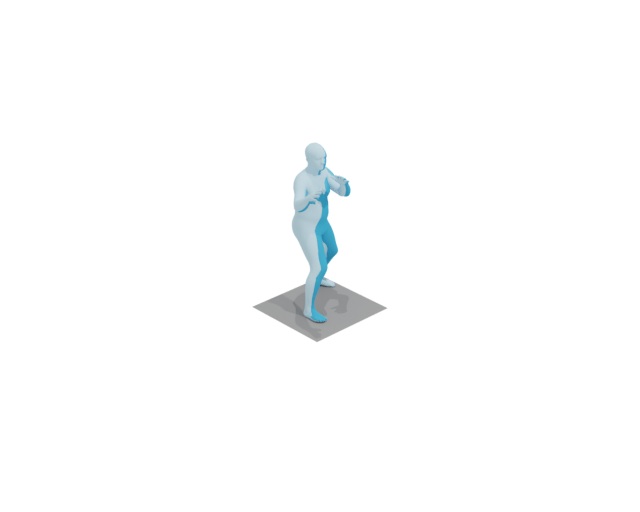}
\includegraphics[width=0.1\linewidth,trim=190 150 190 80,clip]{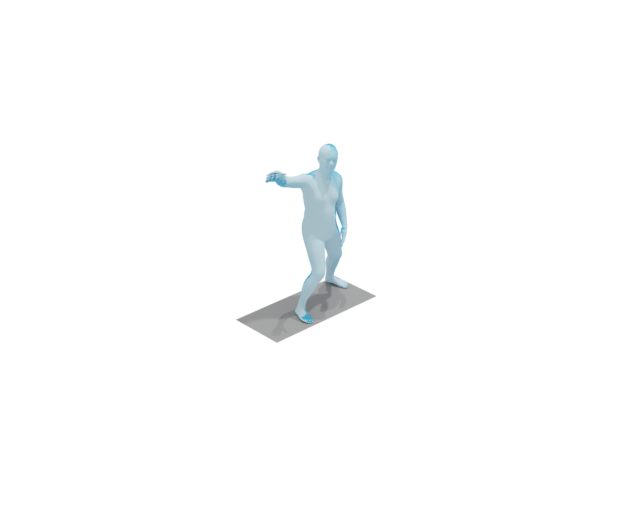}
\includegraphics[width=0.1\linewidth,trim=190 150 190 80,clip]{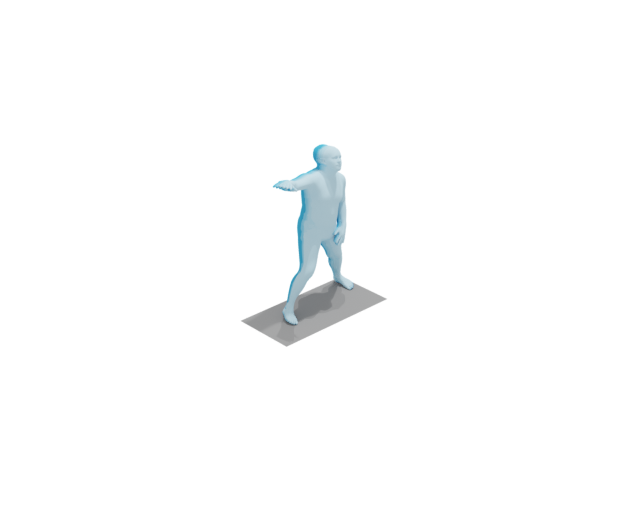}
\includegraphics[width=0.1\linewidth,trim=190 150 190 80,clip]{figs/use_case/436_standing_with_knees_bent_down_with_right_arm_extended/2040_171_182_mesh.png}
\includegraphics[width=0.1\linewidth,trim=190 150 190 80,clip]{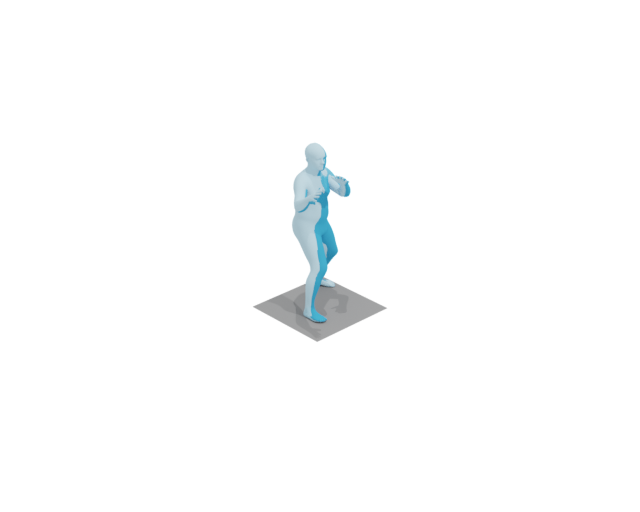}
\includegraphics[width=0.1\linewidth,trim=190 150 190 80,clip]{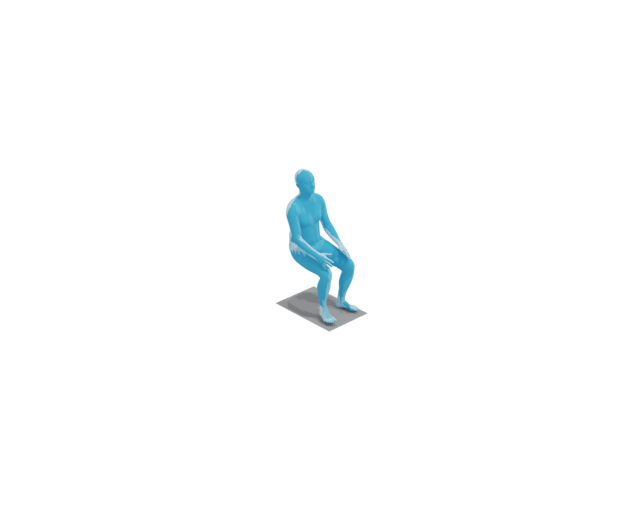}
\includegraphics[width=0.1\linewidth,trim=190 150 190 80,clip]{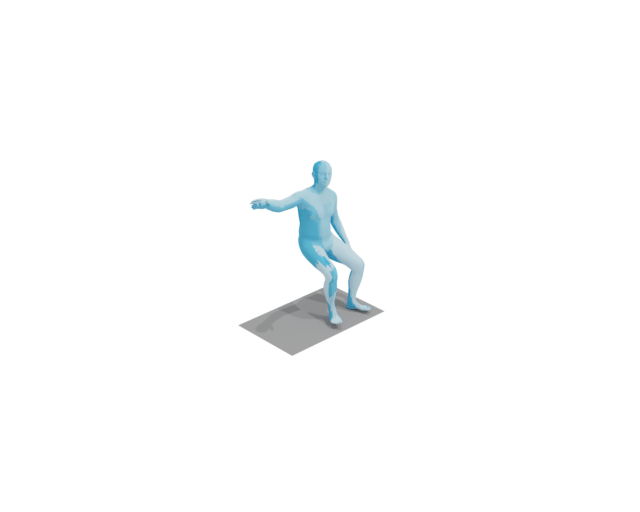}
\includegraphics[width=0.1\linewidth,trim=190 150 190 80,clip]{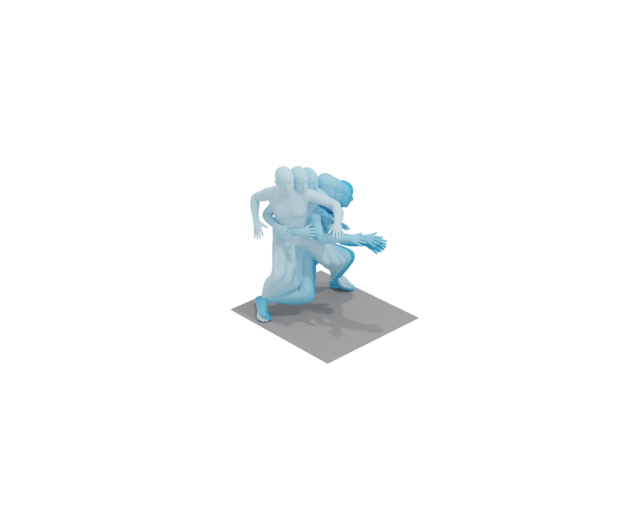}
\includegraphics[width=0.1\linewidth,trim=190 150 190 80,clip]{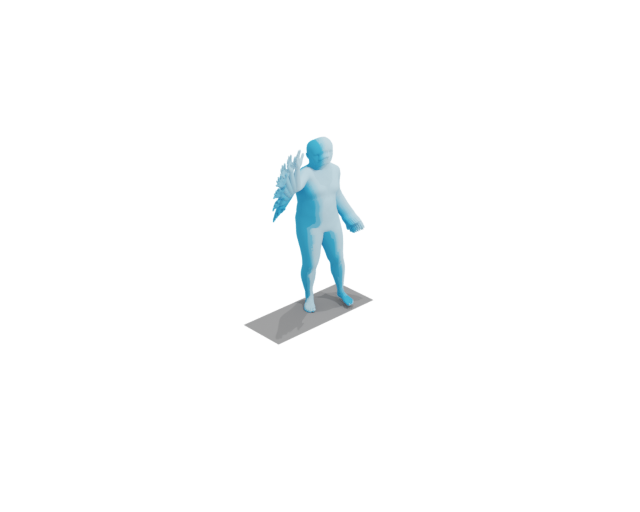}} \vspace{2pt}
\\ \hline
{\Large \textit{run forward}} & 
\raisebox{-.4\height}{\includegraphics[width=0.1\linewidth,trim=190 150 190 80,clip]{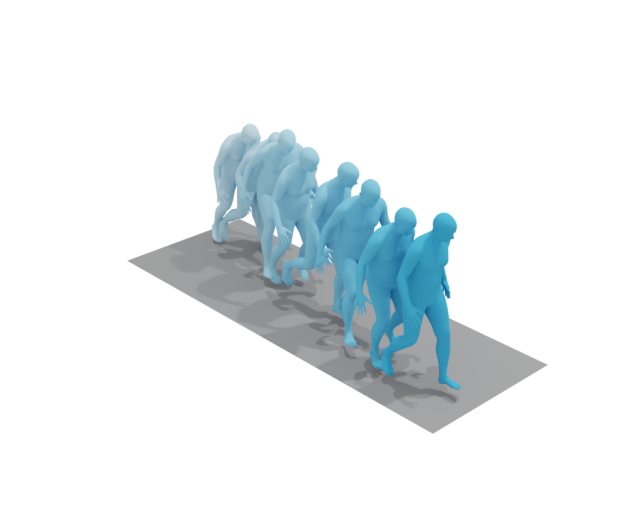}}
&
\raisebox{-.4\height}{
 \includegraphics[width=0.1\linewidth,trim=190 150 190 80,clip]{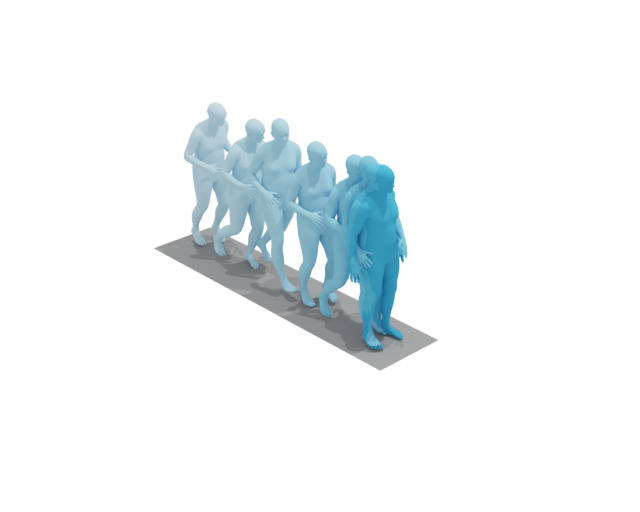}
\includegraphics[width=0.1\linewidth,trim=190 150 190 80,clip]{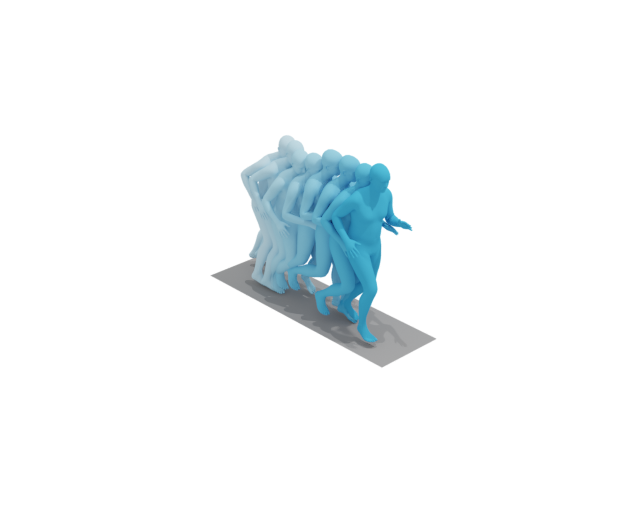}
\includegraphics[width=0.1\linewidth,trim=190 150 190 80,clip]{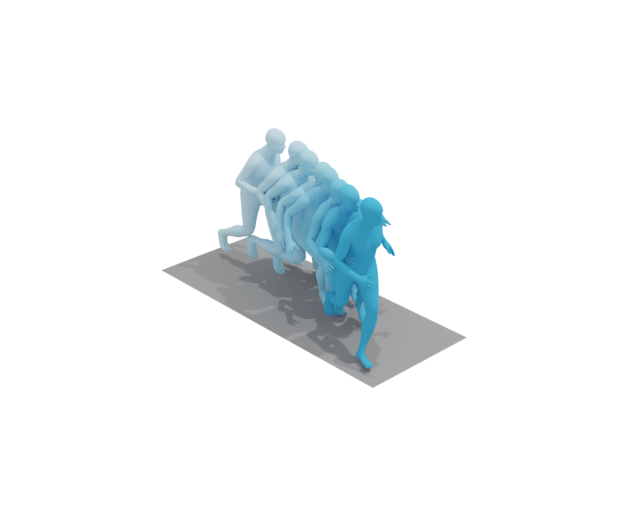}
\includegraphics[width=0.1\linewidth,trim=190 150 190 80,clip]{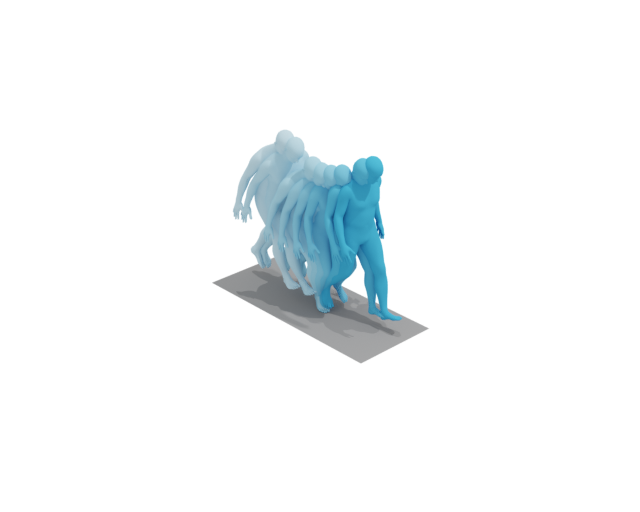}
\includegraphics[width=0.1\linewidth,trim=190 150 190 80,clip]{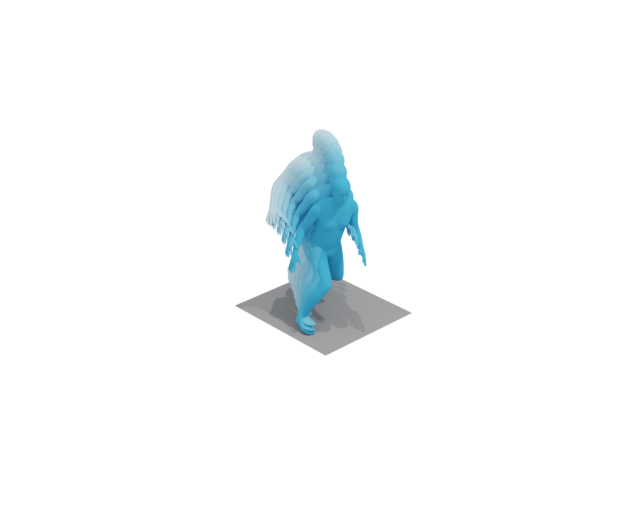}
\includegraphics[width=0.1\linewidth,trim=190 150 190 80,clip]{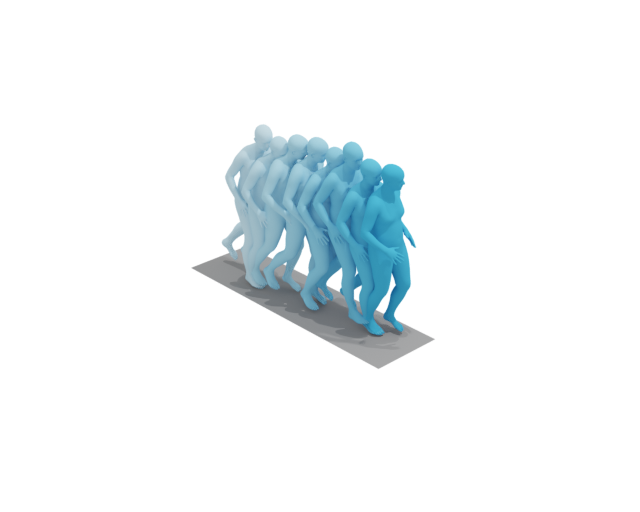}
\includegraphics[width=0.1\linewidth,trim=190 150 190 80,clip]{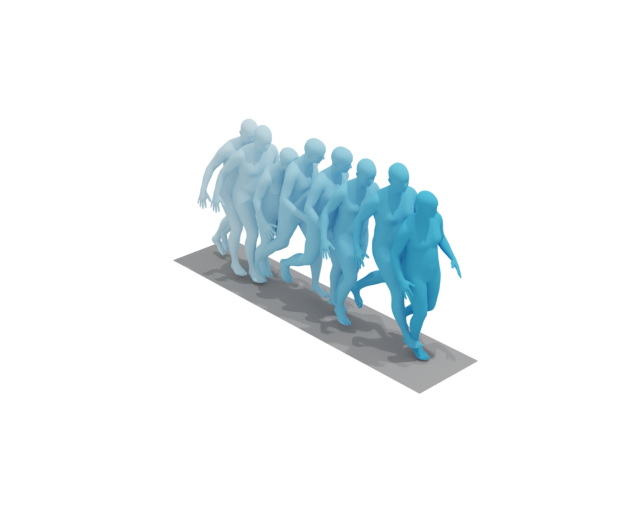}
\includegraphics[width=0.1\linewidth,trim=190 150 190 80,clip]{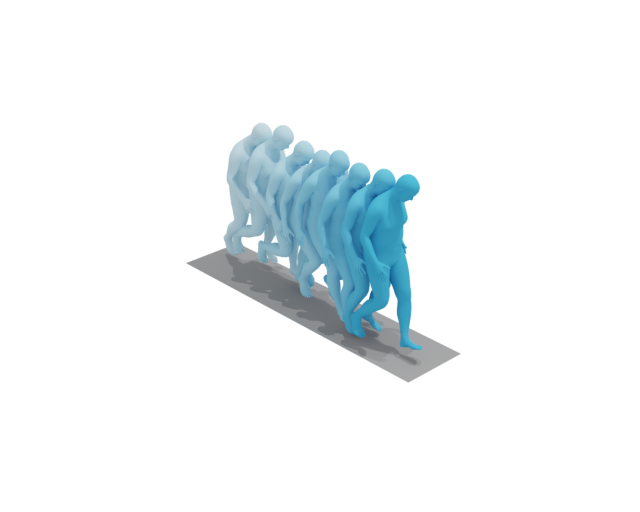}
\includegraphics[width=0.1\linewidth,trim=190 150 190 80,clip]{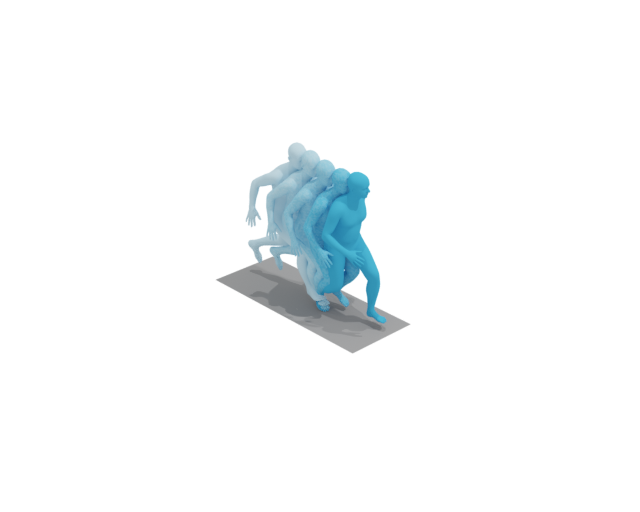}
\includegraphics[width=0.1\linewidth,trim=190 150 190 80,clip]{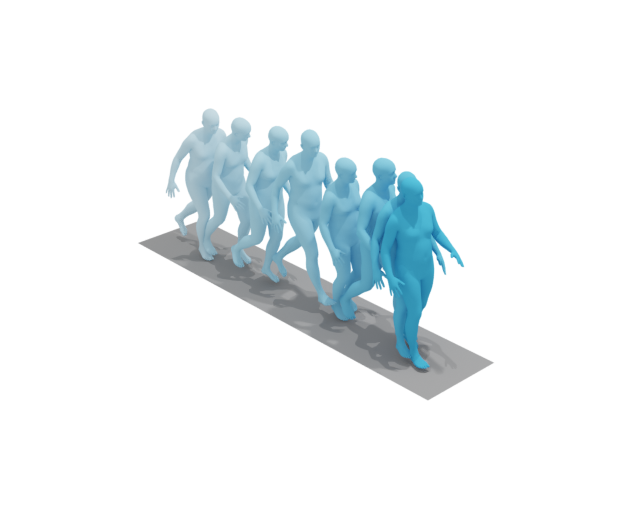}} \vspace{2pt}
\\ \hline
 {\Large \textit{hit something with right hand}} & 
\raisebox{-.4\height}{\includegraphics[width=0.1\linewidth,trim=190 150 190 80,clip]{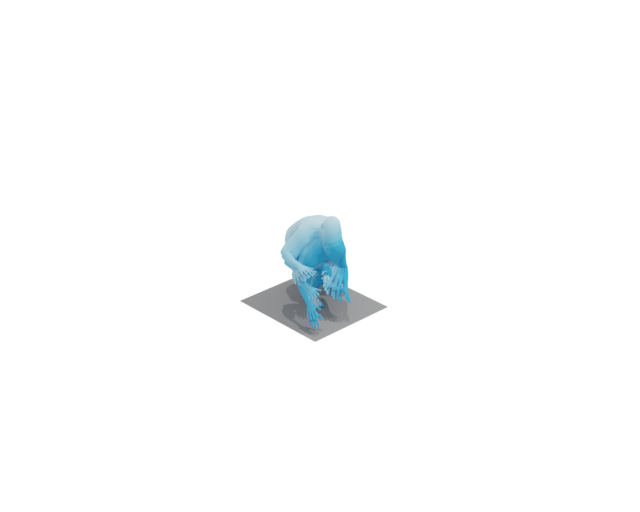}}
&
\raisebox{-.4\height}{
 \includegraphics[width=0.1\linewidth,trim=190 150 190 80,clip]{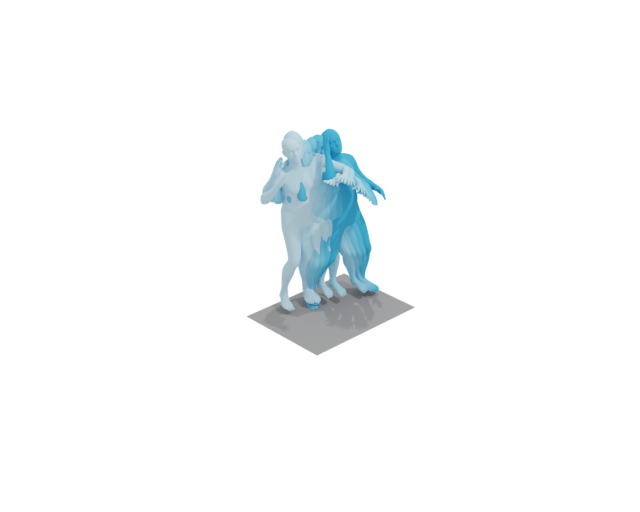}
\includegraphics[width=0.1\linewidth,trim=190 150 190 80,clip]{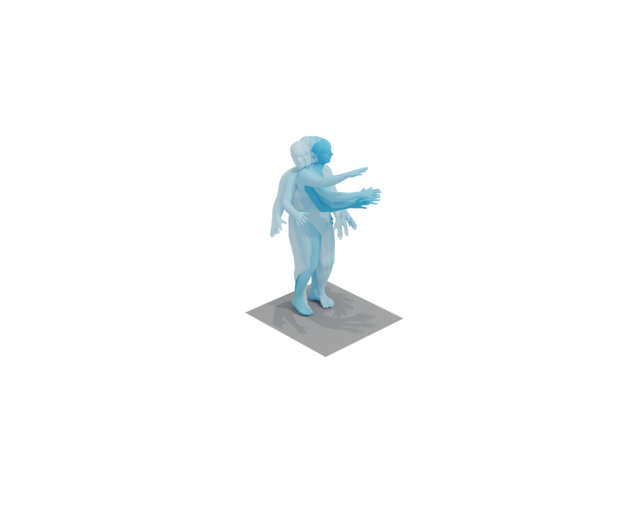}
\includegraphics[width=0.1\linewidth,trim=190 150 190 80,clip]{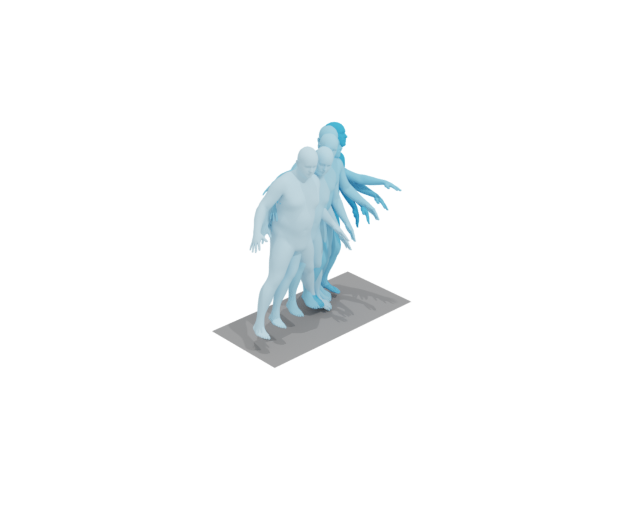}
\includegraphics[width=0.1\linewidth,trim=190 150 190 80,clip]{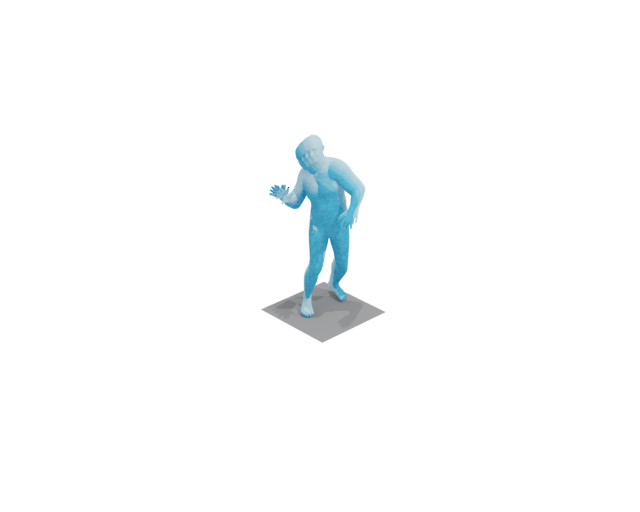}
\includegraphics[width=0.1\linewidth,trim=190 150 190 80,clip]{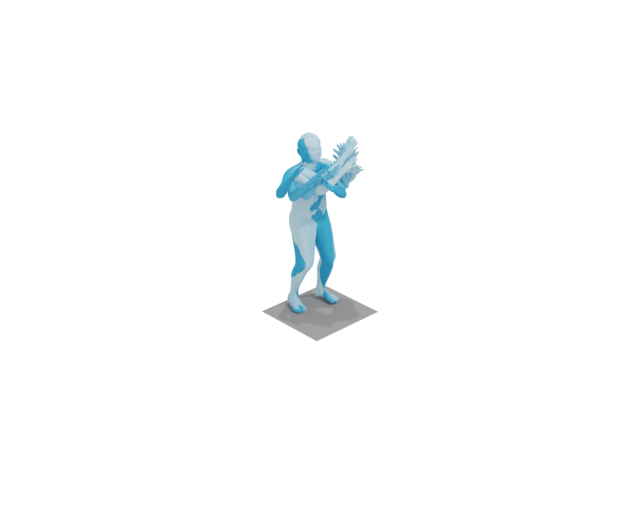}
\includegraphics[width=0.1\linewidth,trim=190 150 190 80,clip]{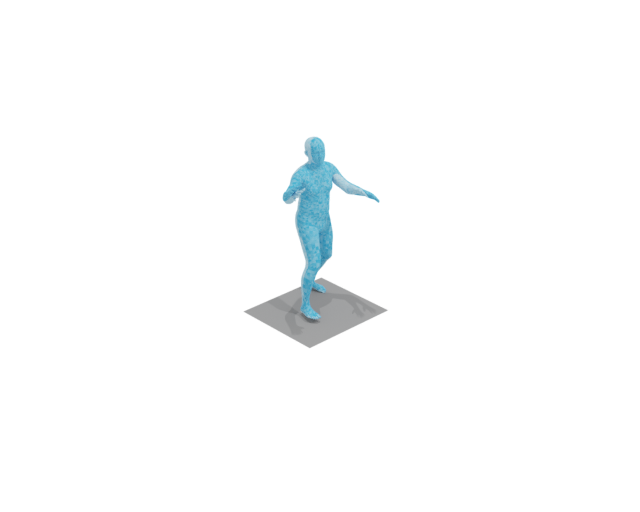}
\includegraphics[width=0.1\linewidth,trim=190 150 190 80,clip]{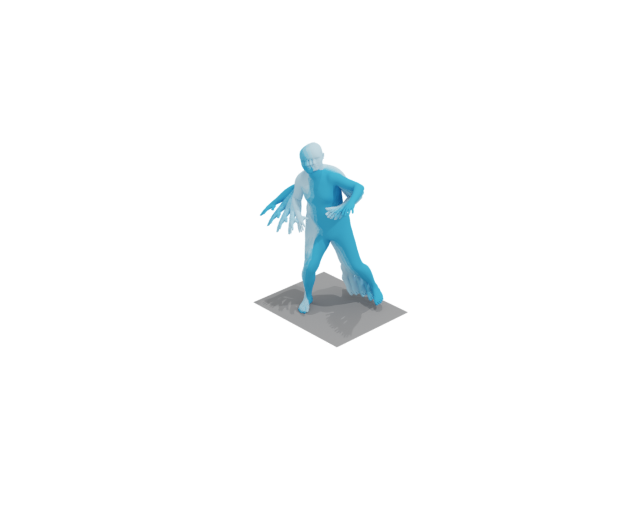}
\includegraphics[width=0.1\linewidth,trim=190 150 190 80,clip]{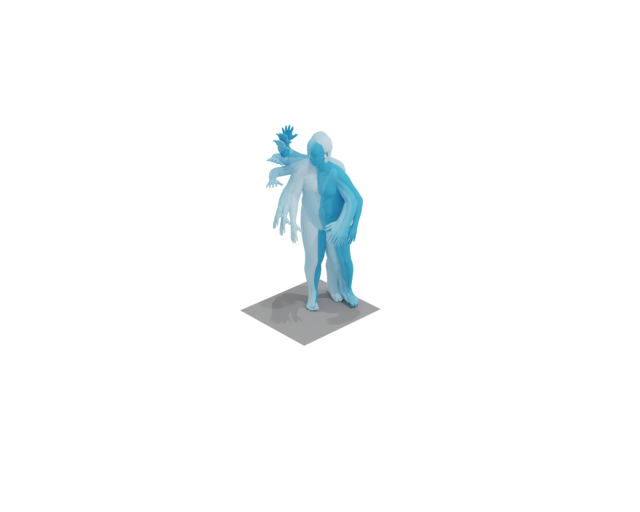}
\includegraphics[width=0.1\linewidth,trim=190 150 190 80,clip]{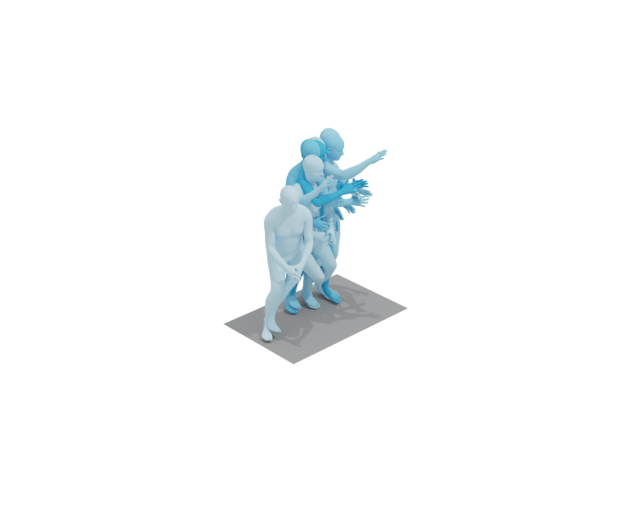}
\includegraphics[width=0.1\linewidth,trim=190 150 190 80,clip]{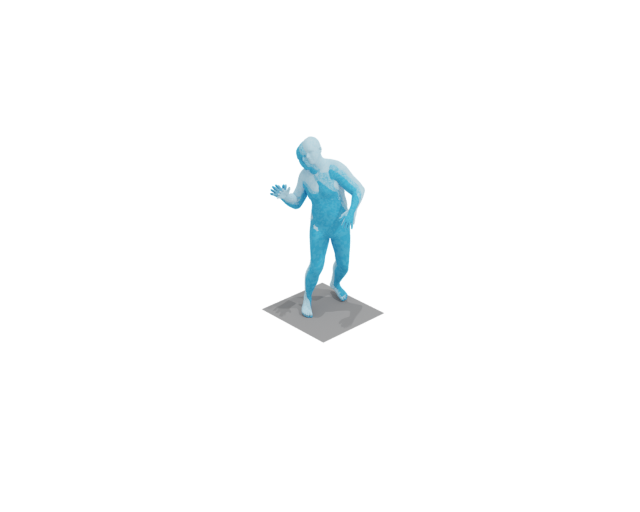}} \vspace{2pt}
\\ \hline 
{\Large \textit{kung fu pose}} & 
\raisebox{-.4\height}{\includegraphics[width=0.1\linewidth,trim=190 150 190 80,clip]{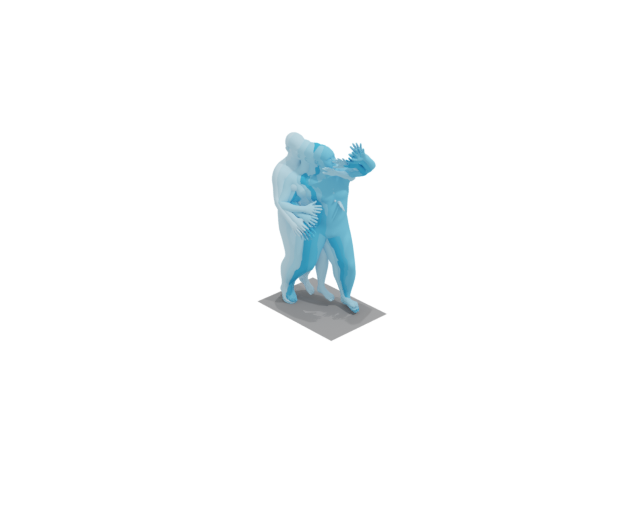}}
&
\raisebox{-.4\height}{
 \includegraphics[width=0.1\linewidth,trim=190 150 190 80,clip]{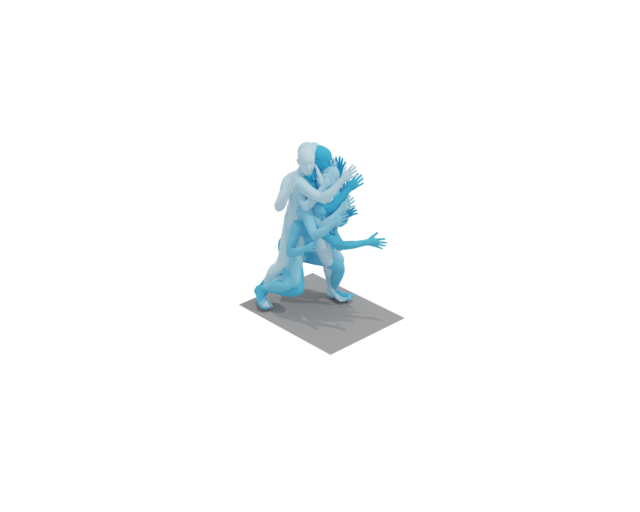}
\includegraphics[width=0.1\linewidth,trim=190 150 190 80,clip]{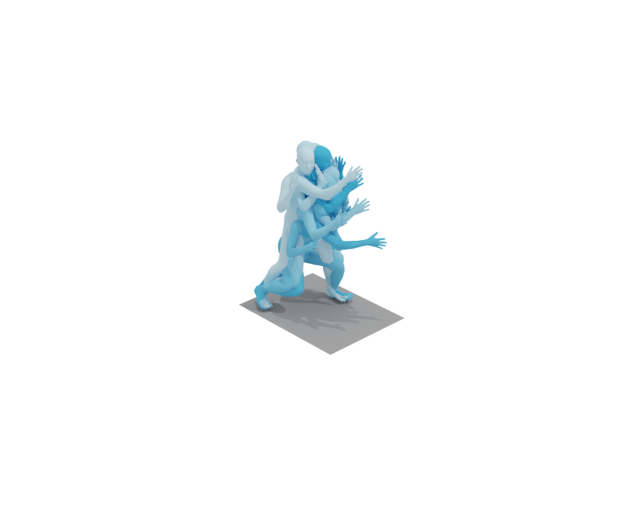}
\includegraphics[width=0.1\linewidth,trim=190 150 190 80,clip]{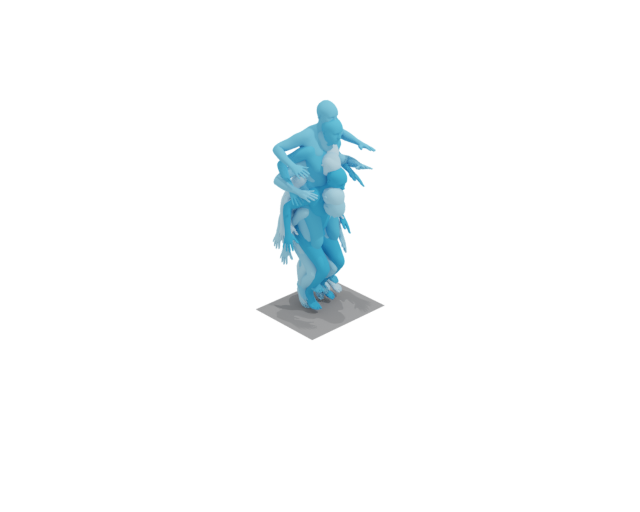}
\includegraphics[width=0.1\linewidth,trim=190 150 190 80,clip]{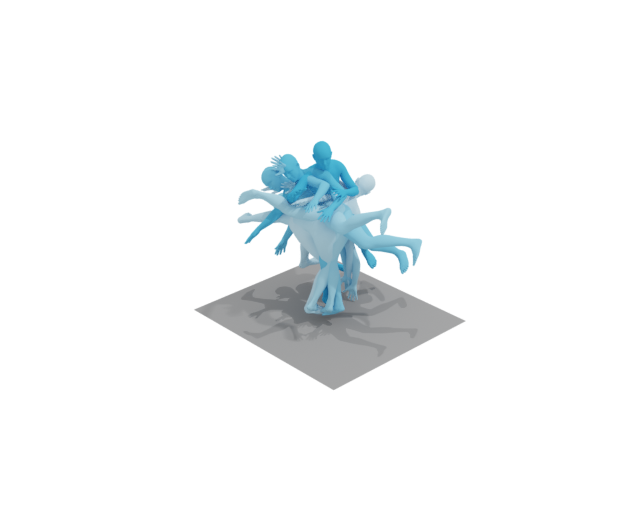}
\includegraphics[width=0.1\linewidth,trim=190 150 190 80,clip]{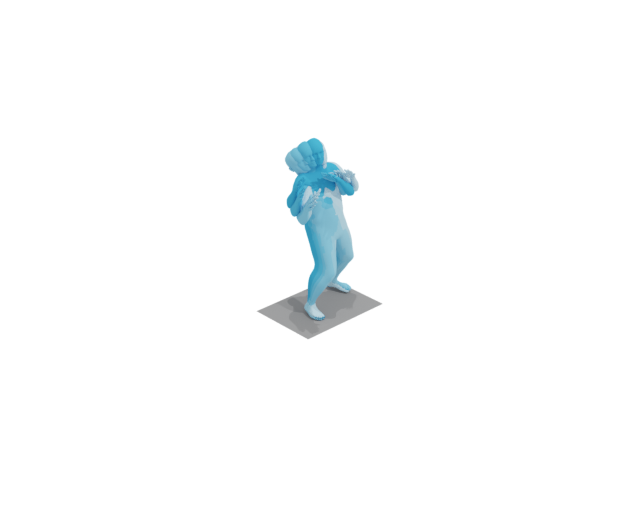}
\includegraphics[width=0.1\linewidth,trim=190 150 190 80,clip]{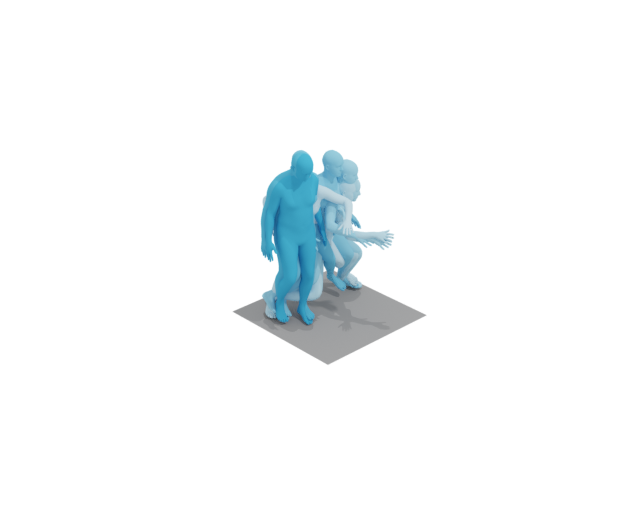}
\includegraphics[width=0.1\linewidth,trim=190 150 190 80,clip]{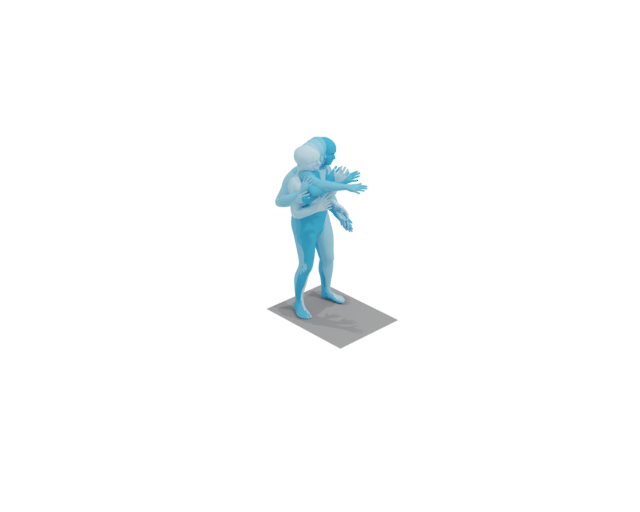}
\includegraphics[width=0.1\linewidth,trim=190 150 190 80,clip]{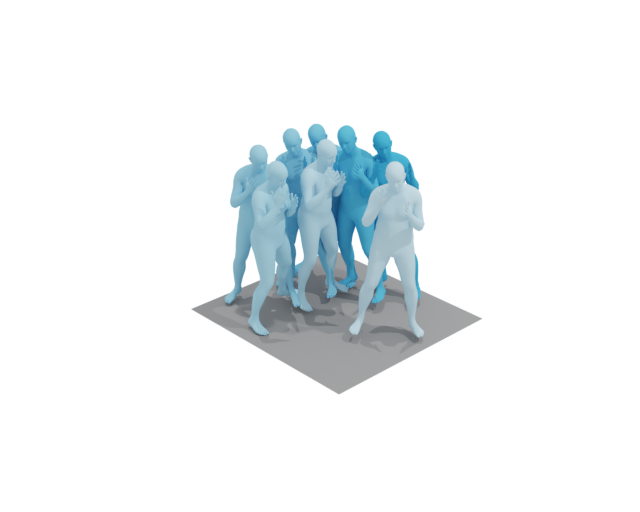}
\includegraphics[width=0.1\linewidth,trim=190 150 190 80,clip]{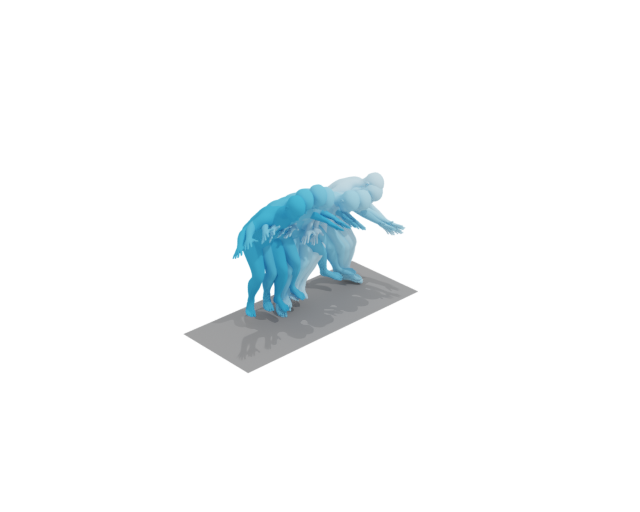}
\includegraphics[width=0.1\linewidth,trim=190 150 190 80,clip]{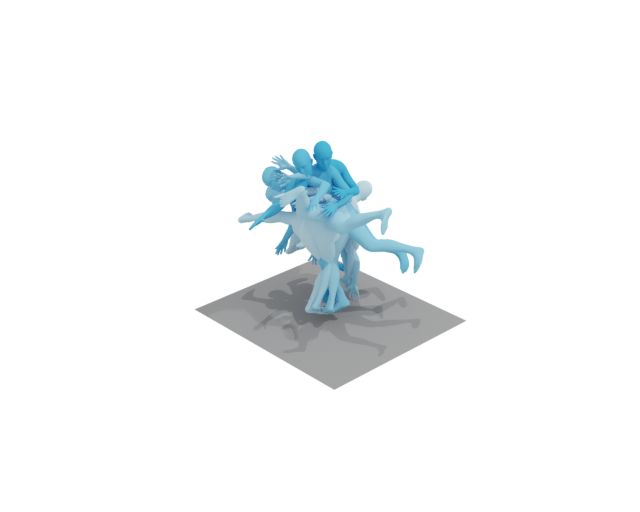}} \vspace{2pt}
\\
\end{tabular}%
}
\caption{\textbf{Use case: corpus-level moment retrieval}. Visualized based on $k=200,\text{ } \lambda=5$. The queries of the BABEL are mainly short sentences. For the retrieved moments, we provide the ground-truth (GT) moments in the second column for qualitative comparison.}
\label{tab:use_case}
\vspace{-10pt}
\end{table*}

\noindent\textbf{Results analysis}. Tables \ref{tab:benchmark_babel} and \ref{tab:benchmark_h3d} summarize the results on two datasets. We report the best and the second-best results with \textbf{bold} and \underline{underlined} font, respectively. First, recent retrieval models, although extended to TSLM with some localization capability, exhibit lower accuracy at high IoU compared to conventional TSLV methods due to limitations of temporal pyramid approaches. Additionally, their evaluation is time-consuming due to the need for sliding windows across multiple scales on all sequences. In contrast, re-implemented TSLV methods offer a faster response. 
Anchor-based or 2D-Map anchor-based MMN shows feasibility in TSLM, but it also suffers from poor performance at high IoU.
Furthermore, on BABEL, regression-based LGI performs even lower than retrieval models. We analyze that during regression optimization, it may be jumping around on the time axis because of an excessive number of false-negative moments.
Finally, {\ul our \tesla{} performs comparably or more weaker than other baselines as they overlook the characteristics of human motion (\eg, \tesla{} vs. MS-DETR $\to$ 29.12 vs. 35.90 IoU@0.7 on HumanML3D (Restore), protocol (a)). By injecting prior knowledge into the model, our \teslams{} outperforms baselines at high IoU}, \eg, surpassing MMN by nearly 4 points at IoU@0.9 (BABEL, protocol (b)). Furthermore, \teslam{} with spatial feature extraction (S-Enc) improves by 1-2 points, indicating that our model better captures the fine-grained nature of motions. Our significant improvements over state-of-the-art can be attributed to \textbf{B)} and \textbf{C)}. Next, we conduct ablations on these components.

\begin{table*}[]
	\centering

 \begin{minipage}[t]{0.03\linewidth}\end{minipage}
	\begin{minipage}[t]{0.3\linewidth}
		\centering
            \subfloat[          \label{tab:ablation:position}]{
			\resizebox{\linewidth}{!}{%
			\centering
			\begin{tabular}{@{}ccc@{}}
                \toprule
                \multicolumn{1}{c|}{\textbf{Position Encoding}} & \multirow{2}{*}{IoU@0.7} & \multirow{2}{*}{mIoU} \\
                \multicolumn{1}{c|}{$\mathcal{E}_{pos}$}  &  &  \\ \midrule
                \multicolumn{1}{c|}{None} & 39.07  & 47.72 \\
                \multicolumn{1}{c|}{Before encoding} &  36.81 & 46.39 \\
                \multicolumn{1}{c|}{After encoding} & \textbf{39.56} &  \textbf{48.57} \\ \bottomrule
                \end{tabular}%
                }
		}

	\end{minipage}
 \begin{minipage}[t]{0.3\linewidth}
		\centering		
		\subfloat[ \label{tab:ablation:order}]{
                \resizebox{\linewidth}{!}{%
			\centering
			\begin{tabular}{@{}ccc@{}}
                \toprule
                \multicolumn{1}{c|}{\textbf{Perturb Rate}} & \multirow{2}{*}{IoU@0.7} & \multirow{2}{*}{mIoU} \\
                \multicolumn{1}{c|}{$\alpha$} &  & \\ \midrule
                \multicolumn{1}{c|}{0.0 (w/o Perturbation strategy)} & 8.57 & 21.35  \\ \midrule
                \multicolumn{1}{c|}{0.3} & 36.74 & 46.41 \\
                \multicolumn{1}{c|}{0.6} & 38.97 & 48.20 \\
                \multicolumn{1}{c|}{0.8} & \textbf{39.56} & \textbf{48.57}  \\
                \multicolumn{1}{c|}{0.9} & 37.26 & 46.72 \\
                \multicolumn{1}{c|}{1.0} & 35.86 & 45.24 \\ \bottomrule
                \end{tabular}%
                }
		}
	\end{minipage}
 \begin{minipage}[t]{0.03\linewidth}\end{minipage}
	\begin{minipage}[t]{0.3\linewidth}
		\centering
		\subfloat[
            \label{tab:ablation:att_cal}]{
			\resizebox{\linewidth}{!}{%
			\centering
			\begin{tabular}{@{}ccc@{}}
                \toprule
                \multicolumn{1}{c|}{\textbf{Components}} & \multirow{2}{*}{IoU@0.7} & \multirow{2}{*}{mIoU} \\
                \multicolumn{1}{c|}{LP-Predictor} &   &  \\ \midrule
                \multicolumn{1}{l|}{MLP-S} & \textbf{39.56} &  \textbf{48.57} \\
                \multicolumn{1}{l|}{\text{    } w/o Dual-alignment}  & 37.74 & 48.21 \\
                \multicolumn{1}{l|}{\text{    } w/o Recovering part}  & 36.77 & 47.22 \\ \bottomrule
                \end{tabular}%
                }
		}
	\end{minipage}
	\vspace{-5pt}
	\caption{\textbf{In-depth analysis on MLP-S}. (a) Study the impact of various perturb-rates in our Perturbation strategy. (b) Exploring the location of positional encoding in TSLM. (c) Detailed ablation of our LP-Predictor.}
 \vspace{-10pt}
	\label{tab:ablation_indepth}
\end{table*}

\begin{table}[]
\centering
\small
\resizebox{\columnwidth}{!}{%
\begin{tabular}{lcccc}
\toprule
\multicolumn{1}{l|}{Method} & IoU@0.5 & IoU@0.7 & IoU@0.9 & mIoU \\ \midrule
\multicolumn{1}{l|}{$\text{MS-DETR}^{\dagger}$~\cite{wang2023ms}} & 53.50 & 36.78 & 12.14  & 48.85 \\
\multicolumn{1}{l|}{MLPBase} & 42.42  & 31.76 &  13.96 &  42.97 \\
\multicolumn{1}{l|}{MLP-S} & 51.12  & 39.56 &  17.10 &  48.57 \\ \midrule
\multicolumn{1}{l|}{MLPBase w/ Glove} & 44.77  & 33.90 & 15.05  &  44.90 \\ 
\multicolumn{1}{l|}{MLP-S w/ Glove} & \textbf{54.17} &  \textbf{40.93} &  \textbf{17.40} &  \textbf{50.96} \\  \bottomrule
\end{tabular}
}
\vspace{-5pt}
\caption{\textbf{Comparison of different word embeddings}. ${\dagger}$ means our re-implementation.}
\label{tab:glove}
\vspace{-0.3cm}
\end{table}

\subsection{Ablation Study \& In-depth Analysis} \label{subsec:ablation}

The following experimental results are reported according to evaluation protocol ``(b) Assigned" on the BABEL dataset.

\noindent\textbf{Main ablation studies.} As shown in Table.~\ref{tab:ablation_main}, we present the contributions of each component in \teslams{} and compare them with \tesla{}. From the results, it can be observed that injecting prior knowledge into the model to assist in match or localization is effective for TSLM.

\noindent\textbf{When to perform position encoding?} Table.~\ref{tab:ablation_indepth}(a) highlights the significance of positional encoding (PE, $\mathcal{E}_{\text{pos}}$). Unlike baseline models adding PE before encoding, we apply PE after encoding, \ie, the sequence matching phase (Sec.~\ref{subsubsec:lpmatcher}). Results suggest that applying PE after encoding is more suitable for TSLM. This might be related to the representation of human motion itself, as each pose is composed of only a few joints. Applying PE before encoding could potentially disrupt the inherent structure of the poses.

\noindent\textbf{The impact of our LP-Matcher.} We varied the perturb-rate $\alpha$ to assess its impact on \teslams{}. In Table.~\ref{tab:ablation_indepth}(b), experiments with various $\alpha$ values show that $\alpha=0$, \ie, w/o Perturbation strategy, makes the model overly reliant on prior knowledge, resulting in a loss of localization ability during inference (mIoU only 21.35). 
The best results are observed at $\alpha=0.8$, indicating that a higher perturbed sequence effectively mitigates training-inference discrepancy. Moreover, Fig.~\ref{fig:avg_higghlight_score} depicts the average highlighting scores of foreground regions for \tesla{} and \teslams{} at different IoU thresholds. The figure shows that \teslams{}$>$\tesla{} and higher IoU values correspond to larger highlighting scores, In other words, having higher highlighting scores in the foreground region increases the chances of accurately locating the target moments.

\noindent\textbf{The impact of our LP-Predictor.} Table \ref{tab:ablation_indepth}(c) presents the results of w/o Dual-alignment and w/o Recovering part. \teslams{} demonstrates a significant improvement over the scenario without prior knowledge injection (\ie, w/o Recovering part), 39.56 vs. 36.77 at IoU@0.7. Furthermore, when w/o Dual-alignment, there is a minor decrease but still higher than w/o the Recovering part. This indicates that conducting recovery learning can further enhance the robustness of the span predictor.

\noindent\textbf{Ours with Glove word embedding}. We present two versions of \tesla{} and \teslams{} in Table \ref{tab:glove} because some baselines in Tables \ref{tab:benchmark_babel} and \ref{tab:benchmark_h3d} were constructed using Glove word embeddings \cite{pennington2014glove}, while our \teslam{} is based on RoBERTa. According to the results, the TSLM final performance is not significantly affected by the use of RoBERTa without fine-tuning. Therefore, we consider the comparison of \teslams{} with TSLV sota methods to be fair.

\begin{figure}[t]
\centering
\vspace{-5pt}
\includegraphics[width=1.0\linewidth]{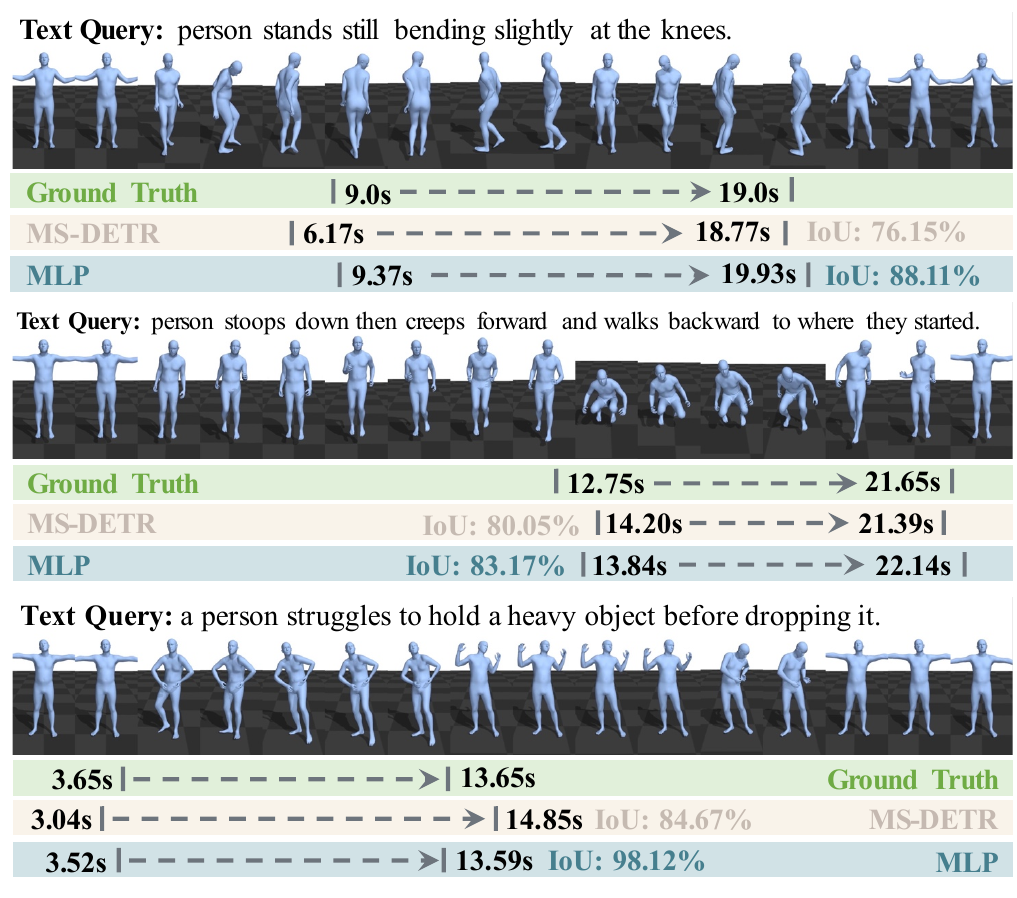}
\vspace{-20pt}
\caption{Visualization of the ground-truth moment and predictions by competitors on the HumanML3D (Restore) dataset.}
\label{fig:vesus}
\vspace{-0.3cm}
\end{figure}

\subsection{Qualitative Results} \label{subsec:qualitative}

In Figs.~\ref{fig:qualitative_success2} and~\ref{fig:qualitative_success1}, we present two successful cases on BABEL and HumanML3D (Restore), respectively, denoted as 1), 2), 3), and 4) sequentially. From 1) and 2), it is evident that within the foreground regions (highlighted in yellow), \teslam{} generally exhibits higher highlighting scores $\mathcal{S}_{\text{LP}}$ than scores $\mathcal{S}_{\text{SM}}$ of \tesla{}, and in the background region, $\mathcal{S}_{\text{LP}}$ are generally lower than $\mathcal{S}_{\text{SM}}$, as evident in the red rectangle in 1). This favors confining the predicted start/end boundaries within the ground-truth region, demonstrating the contribution of the LP-Matcher. In 3) and 4), we visualize the probability distribution of the recovering part in LP-Predictor. Please focus on the red rectangle in 3), where it can be seen that compared to $\mathcal{P}_{\text{s}}$ of \tesla{}, $\mathcal{P}_{\text{s}}$ of \teslam{} is closer to $\mathcal{P}_{\text{s}}^{\text{rec}}$.

\begin{figure*}[htbp]
\centering
\vspace{-5pt}
\includegraphics[width=0.99\linewidth]{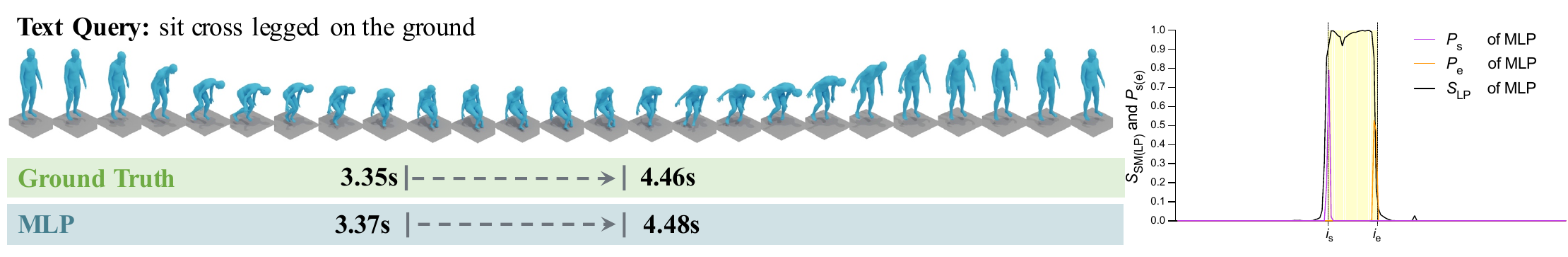}
\includegraphics[width=0.99\linewidth]{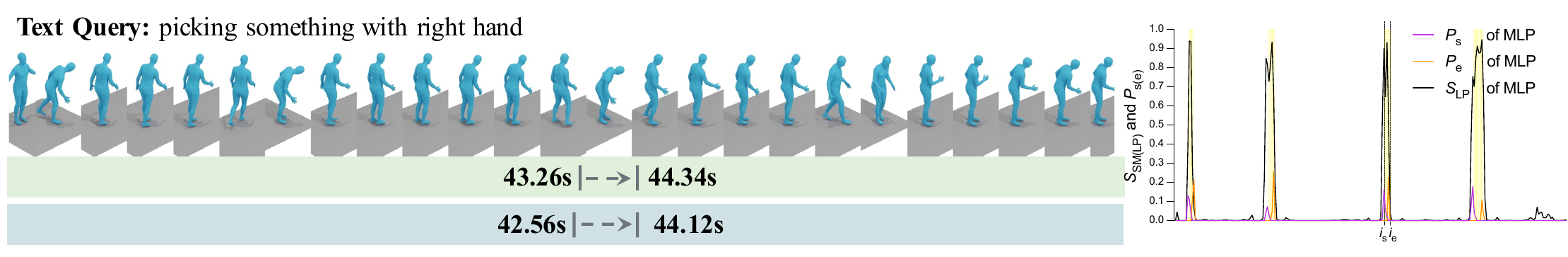}
\includegraphics[width=0.99\linewidth]{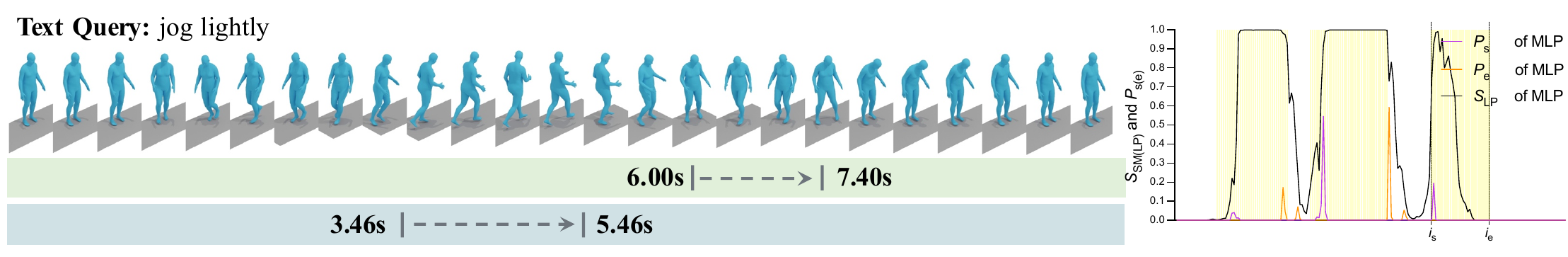}
\vspace{-2pt}
\caption{More qualitatively successful grounding cases on BABEL. The assigned ``false negative moments" are also highlighted in yellow. In the third result, \teslam{} successfully locates the ``false negative moment", which corresponds to the action of ``jog lightly".}
\vspace{-0.35cm}
\label{fig:qualitative_success3}
\end{figure*}

Fig. \ref{fig:vesus} illustrates a qualitative comparison between our \teslam{} and MS-DETR \cite{wang2023ms} on the HumanML3D (Restore) dataset. It can be observed that our \teslam{} achieves more accurate localization.  In addition, Fig. \ref{fig:qualitative_success3} visualizes more successful localization cases on BABEL, where ``false-negative moments" are also highlighted with a yellow background.

We examined the failure cases on BABEL depicted in Fig.~\ref{fig:qualitative_failure}, labeled as (i) and (ii). In case (i), \teslam{} localized a moment of a person standing up for the query ``nod head". This could be because nodding might occur simultaneously with standing up, and it could be difficult to discern the subtle head movement that goes along with nodding. In case (ii), the term ``moving sideways" may not have been fully understood by \teslam{} because it localized a moment with the semantics of a ``u-pose".

\section{Use case: Corpus-level Moment Retrieval} \label{subsec:use_case}

While our focus is moment retrieval (localization) within single-motion sequences, the trained model can be flexibly applied to various use cases. Here, we qualitatively assess the limits of our approach by collaboratively coordinating with the lightweight motion-text retrieval model, Rehamot \cite{yan2023cross}, to perform moment retrieval tasks across the entire BABEL corpus. This is more in line with real-world scenarios.

\noindent\textbf{Implementation details.} Corpus-level moment retrieval (CMR) consists of two steps: (1) retrieving motions in the corpus relevant to the query, and (2) locating moments within the retrieved motions. This is fundamentally similar to  video corpus moment retrieval tasks~\cite{lei2020tvr, zhang2023video}. Specifically, given a query and a corpus containing $M$ motions, we use Rehamot to compute the similarity between the query and each chunk of each motion (chunk size fixed at 10s with 90\% overlap between chunks). We select the maximum similarity as the retrieval score for the query-motion pair, leading to $\mathbf{r}=[r_{1},r_{2},\ldots,r_{M}]$. For the top-$k$ motions retrieved based on $\mathbf{r}$, we employ \teslam{} to calculate localization scores $\text{p}^{se}$ for certain candidate moments (Eq.~\eqref{eq:infer}). The final CMR score is computed as $\text{p}^{se} \times \text{exp}(\lambda \cdot r)$, where the exponential term and hyperparameter $\lambda$ are used to balance the importance of retrieval and localization scores.

\noindent\textbf{Qualitative results.} Table.~\ref{tab:use_case} visualizes the top-10 CMR results for the queries, providing the ground-truth (GT) moments as a reference. The results demonstrate a good semantic representation of the queries. For instance, in the case of ``kung fu pose," although the retrieval results may not be similar to the GT, they exhibit other manifestations of ``kung fu pose" perceptually.

\noindent\textbf{Quantitative results.}
We consider two evaluation metrics to assess CMR:
(a) \textit{Correct-recall}. ``$\textrm{Recall@}n, \textrm{IoU}@\mu$'' is used to evaluate the quality of correct recalls based on the sorted CMR score list.
(b) \textit{Relevance-based}. As demonstrated in the qualitative results, there are many retrieved moments that reflect the query semantics but are not near the GT. Metric (a) overly penalizes such cases. Therefore, we adopt the commonly used metric in recommender systems, namely, Discounted Cumulative Gain (DCG). ``$\textrm{DCG@}n$" evaluates the relevance of specific queries by looking at the top $n$ positions in the CMR list. DCG assumes that moments with higher relevance appearing later in the list should be penalized using a logarithmic factor: $\textrm{DCG@}n=\sum_{i=1}^{n} \frac{rel_{i}}{\text{log}_{2}(i+1)}$. Here, we define two \textit{relevance} functions to compute $rel$:
(i) \texttt{Text-Motion} relevance. Calculating similarity scores between the text query and the retrieved moments using Rehamot.
(ii) \texttt{Text-Text} relevance. A more robust relevance using proxy relevance between texts~\cite{messina2021transformer}. We translate the retrieved moments into text using a well-established motion-to-text model~\cite{guo2022tm2t} Then, similar to Sec.~\ref{subsec:training_strategy}, MPNet~\cite{song2020mpnet} is used to calculate the similarity score. The translated text is obtained through greedy search.

\begin{table}[]
\centering
\resizebox{\columnwidth}{!}{%
\begin{tabular}{@{}l|cccccc@{}}
\toprule
\multirow{3}{*}{\textbf{Hyperparameter}} & \multicolumn{6}{c}{\textbf{Recall@\textit{n}}} \\
 & \multicolumn{3}{c}{IoU@0.5} & \multicolumn{3}{c}{IoU@0.7} \\ \cmidrule(l){2-4} \cmidrule(l){5-7} 
 & $n$=10 & $n$=50 & $n$=100 & $n$=10 & $n$=50 & $n$=100 \\ \midrule
$k=100,\text{ }\lambda=5$ & 3.33 & 7.96 & 10.34 & 2.65 & 6.38 & 8.22 \\ 
$k=200,\text{ }\lambda=5$ & 3.31 & 8.10 & 11.12 & 2.61 & 6.50 & 8.87 \\
$k=100,\text{ }\lambda=10$ & \textbf{3.70} & 8.23 & 10.48 & \textbf{3.01} & 6.57 & 8.39 \\
$k=200,\text{ }\lambda=10$ & 3.69 & \textbf{8.41} & \textbf{11.44} & 2.98 & \textbf{6.75} & \textbf{9.06} \\ \midrule
\multirow{3}{*}{\textbf{Hyperparameter}} & \multicolumn{6}{c}{\textbf{DCG@\textit{n}}} \\
 & \multicolumn{3}{c}{Text-Motion~\cite{yan2023cross}} & \multicolumn{3}{c}{Text-Text~\cite{guo2022tm2t, song2020mpnet}} \\ \cmidrule(l){2-4} \cmidrule(l){5-7} 
 & $n$=10 & $n$=50 & $n$=100 & $n$=10 & $n$=50 & $n$=100 \\ \midrule
\textbf{Ideal DCG} & 4.54 & 12.90 & 20.94 & 4.54 & 12.90 & 20.94 \\ \hline
$k=100,\text{ }\lambda=5$ & 1.66 & 4.48 & 7.11 & 2.64 & 7.35 & 11.83 \\ 
 $k=200,\text{ }\lambda=5$ & 1.60 & 4.25 & 6.74 & 2.61 & 7.28 & 11.71 \\
$k=100,\text{ }\lambda=10$ & \textbf{1.79} & \textbf{4.72} & \textbf{7.47} & \textbf{2.69} & \textbf{7.51} & \textbf{12.08} \\
$k=200,\text{ }\lambda=10$ & 1.78 & 4.63 & 7.23 & 2.67 & 7.41 & 11.89 \\ \bottomrule
\end{tabular}%
}
\vspace{-5pt}
\caption{\textbf{Quantitative results on CMR}. We evaluated the performance of CMR from both recall and relevance aspects. For the latter, we computed the Ideal DCG (where the $rel$ value is always 1.0 at any position) for reference. In the evaluation of text-text relevance, we refrained from using tools like Spacy as they are sensitive to human-centric action descriptions. \eg, using Spacy to calculate the similarity between ``clockwise" and ``counter-clockwise" yields a high similarity of 0.81, which is evidently unreasonable.}
\label{tab:cmr}
\end{table}

We empirically vary hyperparameters $k$ and $\lambda$ to determine the limits of \teslam{}. As shown in Table.~\ref{tab:cmr}, we observe that Rehamot+\teslam{} achieves reasonable results. The discounted cumulative gain reaches around 50\% (\eg, DCG@100 exceeds 10, with an ideal value of 20.94). We believe that CMR can considered as a promising direction for future research.

\section{Conclusion}

In this work, we investigate the unexplored problem of text-to-motion localization. This is a meaningful exploration. Confronting the low contextual richness and the semantic ambiguity between frames inherent in human motion, our proposed \teslam{} achieved state-of-the-art performance in TSLM by injecting label-prior knowledge. Furthermore, collaborating with the retrieval model, \teslam{} showcases robust generalization capabilities in corpus-level moment retrieval.

Future work may consider integrating factorized spatial and temporal feature encoders to construct a spatio-temporal cooperative localization framework, which could bring further benefits.

\bibliographystyle{IEEEtran}
\bibliography{sample-base}

\end{document}